\documentclass[sn-basic]{sn-jnl}% Basic Springer Nature Reference Style/Chemistry Reference Style

\usepackage{graphicx}
\usepackage[T1]{fontenc}
\usepackage{amsmath,amssymb,amsfonts}%
\usepackage[title]{appendix}%
\usepackage{tabularray}%
\usepackage{multirow}%
\usepackage{float}
\usepackage{booktabs,xcolor,graphicx}
\usepackage{adjustbox}
\usepackage{rotating}
\usepackage{array}
\usepackage{tabularx}
\usepackage{makecell}
\usepackage{pdflscape}
\usepackage{rotating}
\usepackage{svg}
\begin{document}

\title[Advances in Diffusion Models for Image Data Augmentation: A Review of Methods, Models, Evaluation Metrics and Future Research Directions]{Advances in Diffusion Models for Image Data Augmentation: A Review of Methods, Models, Evaluation Metrics and Future Research Directions}

\author[1]{\fnm{Panagiotis} \sur{Alimisis}}\email{csi23301@hua.gr}
\author[1]{\fnm{Ioannis} \sur{Mademlis}}\email{imademlis@hua.gr}
\author[2,3]{\fnm{Panagiotis} \sur{Radoglou-Grammatikis}}\email{pradoglou@uowm.gr, pradoglou@k3y.bg}
\author[2]{\fnm{Panagiotis} \sur{Sarigiannidis}}\email{psarigiannidis@uowm.gr}
\author*[1]{\fnm{Georgios Th.} \sur{Papadopoulos}}\email{g.th.papadopoulos@hua.gr}

\affil[1]{\orgdiv{Department of Informatics and Telematics}, \orgname{Harokopio University of Athens}, \orgaddress{\street{Thiseos 70}, \city{Athens}, \postcode{GR 17676}, \state{Attiki}, \country{Greece}}}

\affil[2]{\orgdiv{Department of Electrical and Computer Engineering}, \orgname{University of Western Macedonia}, \orgaddress{\street{Active Urban Planning Zone}, \city{Kozani}, \postcode{GR 50150}, \state{Kozani}, \country{Greece}}}

\affil[3]{\orgdiv{K3Y}, \orgaddress{\street{Studentski district, Vitosha quarter, bl. 9}, \city{Sofia}, \postcode{BG 1700}, \state{Sofia City Province}, \country{Bulgaria}}}

\abstract{Image data augmentation constitutes a critical methodology in modern computer vision tasks, since it can facilitate towards enhancing the diversity and quality of training datasets; thereby, improving the performance and robustness of machine learning models in downstream tasks. In parallel, augmentation approaches can also be used for editing/modifying a given image in a context- and semantics-aware way. Diffusion Models (DMs), which comprise one of the most recent and highly promising classes of methods in the field of generative Artificial Intelligence (AI), have emerged as a powerful tool for image data augmentation, capable of generating realistic and diverse images by learning the underlying data distribution. The current study realizes a systematic, comprehensive and in-depth review of DM-based approaches for image augmentation, covering a wide range of strategies, tasks and applications. In particular, a comprehensive analysis of the fundamental principles, model architectures and training strategies of DMs is initially performed. Subsequently, a taxonomy of the relevant image augmentation methods is introduced, focusing on techniques regarding semantic manipulation, personalization and adaptation, and application-specific augmentation tasks. Then, performance assessment methodologies and respective evaluation metrics are analyzed. Finally, current challenges and future research directions in the field are discussed.}


\maketitle

\section{Introduction}
\label{sec:introduction}

Modern computer vision has been dominated by the so-called Deep Learning (DL) paradigm, which relies on the use of large-scale on Deep Neural Networks (DNNs) \cite{chai2021deep}. DNNs have so far exhibited outstanding performance in a wide set of visual understanding tasks. However, this eminent visual interpretation and reasoning capability is accompanied by the increased need for ever larger and sufficiently diverse training datasets \cite{shrestha2019review}. On the other hand, as image analysis tasks become increasingly intricate and demanding, the ability of DNNs to generalize robustly is hindered by limitations in training data quantity, diversity, and potential bias \cite{MUMUNI2022100258}. As a result, data requirements has emerged as a rather prominent topic, since a sufficient volume of training samples is essential for fully harnessing the capabilities of DNNs \cite{zhang2211expanding}. On the contrary, real-world image datasets, especially regarding specific-targeted application domains, often suffer in these aspects, even to the point of containing perfectly correlated training images that are proven to be essentially redundant \cite{yang2022image}.

Image augmentation constitutes a common preprocessing step in machine learning workflows, designed to increase the visual diversity of the training dataset, without introducing additional independent images. The goal is to enhance the learning process, while operating within a fixed set of data \cite{Shorten2019ASO}. Augmentation addresses limitations in dataset size, by automatically creating additional variants of existing training images. These variants are generated by directly modifying the original images, ensuring that the transformed images differ in appearance, but retain their semantic content \cite{xu2023comprehensive}. In contrast, general image generation (or image synthesis) involves sampling from a model that approximates the overall data distribution of the dataset, rather than transforming individual training images. This approach generates entirely new images that are not directly derived from specific originals, but they are instead representative of the underlying characteristics of the whole dataset \cite{cao2024survey}. Extending the dataset with such augmented images, either synthesized or modified, increases its diversity and improves, in many cases, the downstream learning and recognition performance of DNNs that are being trained using it \cite{zhou2023using}. This behaviour stems from image augmentation essentially acting as an additional regularizing mechanism while training the DNN and, as a result, helping to prevent overfitting \cite{perez2017effectiveness, Shorten2019ASO}. 

Traditional approaches for image augmentation, such as geometric transformations (e.g., image rotation, flip, crop, scaling, horizontal/vertical translation, squeezing, etc.) and color space adjustments or photometric transforms (e.g., blurring, sharpening, jittering, etc.) are still very common \cite{Xu_2023, Shorten2019ASO, yang2022image}. Multiple transformations of this type can be composed together, so that an even wider set of augmented images can be generated from the original dataset. These methods leverage domain knowledge to produce synthetic examples similar to the initial ones. More recently proposed image augmentation methods in this general vein are a set of strategies for systematically corrupting the original images, in order to generate augmented variants. This category of methods includes, among others: a) `Mixup' \cite{zhang2017mixup}, which uses convex combinations of pairs of training images and their labels, b) `Cutout' \cite{devries2017improved}, which randomly masks square regions of an input image, c) `Cutmix' \cite{yun2019cutmix}, which randomly combines two training images by masking the first with a region of the second (and vice versa), d) `Patchshuffle' \cite{kang2017patchshuffle}, which uses a kernel filter to randomly swap the pixel values in a sliding window, e) `Copy-Paste' \cite{dwibedi2017cut}, which pastes segmented object instances onto random background images, f) `Co-Mixup' \cite{kim2021co}, which optimizes saliency-guided mixing of multiple inputs while encouraging diversity among outputs, g) `RandAugment' \cite{cubuk2020randaugment}, which randomly selects $N$ transformations from a predefined set, each applied with magnitude $M$, eliminating the need for separate policy search, h) `GuidedMixup' \cite{kang2023guidedmixup}, which uses saliency maps to guide image mixing while preserving salient regions through pixel-wise mixing ratios, and i) `CAL-AUG' \cite{rao2021counterfactual}, which randomly replaces learned attention maps with counterfactual ones.

The effectiveness though of the above-mentioned relatively simple and straightforward augmentation methods is being increasingly challenged by the complexity and variability of contemporary image analysis demands. Although such strategies can be effective in increasing data diversity for simple tasks, they are mostly unable to capture the underlying structure and complex relationships present in high-dimensional image data. Additionally, many of them require domain-specific knowledge and dataset-specific calibration, in order to be applied correctly \cite{wu2023detail}. Moreover, the needs of DNNs for large training datasets and effective regularization are ever-growing, rendering image augmentation a critically important component of modern machine learning \cite{zhang2211expanding}.

Unlike traditional methods, which directly manipulate existing images to generate variants, Diffusion Models (DMs) can be exploited for image augmentation, by learning to synthesize new, realistically-looking and plausible images, given a training dataset \cite{zhang2023sine, xu2024cyclenet, asperti2023image}. These are general image generation algorithms, aiming to learn a model of the underlying distribution of the selected training dataset and then allow its sampling, in order to synthesize novel images. These generated images do not have a direct one-to-one correspondence with the original training ones, but instead reflect the dataset's underlying statistical patterns. Generative methods can be used for various tasks, such as procedural content generation and simulation, but they can also implement dataset expansion, as a specific type of image augmentation at the dataset level. Additionally, when coupled with a conditioning mechanism, they can replace traditional augmentation approaches and synthesize modified variants of specific given images, through conditional generation that allows one-to-one correspondence between an original training image and a synthesized variant. DMs are a sophisticated class of generative DNNs that excel in implicitly modeling the underlying data generating distribution and the structure of complex images. This capability allows them to essentially sample fake novel images from their training dataset's distribution, which are simultaneously diverse, highly realistic and representative of unseen data scenarios, as they encompass subtle details and preserve the inherent structure of the original dataset \cite{zhang2211expanding, trabucco2023effective}. Thus, they can be utilized for meaningfully augmenting the latter.

The learning paradigm of DMs, which relies on iteratively applying noise to the training images and subsequently learning to reverse the process, has shown significant promise in image augmentation, when compared against competing generative models (e.g., Generative Adversarial Networks) \cite{ho2020denoising}. Additionally, recent advancements in DMs enable the conditioning of the image synthesis process via class labels, textual descriptions, or input images \cite{rombach2022highresolution}. This level of user control allows for targeted image augmentation, generating images that fulfill specific requirements based on the task at hand.  

\begin{figure*}
    \centering
    \includegraphics[width=0.8\textwidth]{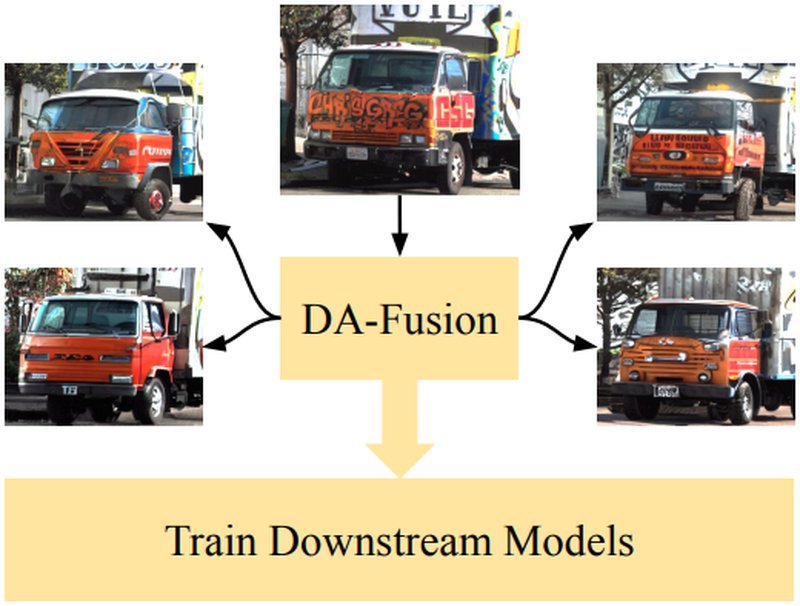}
    \caption{Semantic alteration of truck graffiti design using diffusion models. Image from \cite{trabucco2023effective}.}
    \label{fig:DAFusionTruck}
\end{figure*}

The recent advancement in generative image synthesis through DMs and multimodal strategies (e.g., text-conditioned image creation) has been complemented by the use of large-scale pretraining on massive datasets, in the vein of the Foundation Model (FM) trend \cite{rombach2022highresolution, podell2023sdxl, esser2024scaling, saharia2022photorealistic}. This approach has led to the availability of pretrained DMs that can generate images with natural variations in appearance (e.g., changing the design of the graffiti on a truck, as illustrated in Fig.~\ref{fig:DAFusionTruck}) and, hence, can be directly exploited for sophisticated image augmentation without significant human effort \cite{dunlap2023alia}.  

Despite the increasing importance of DMs in image augmentation, there is a \textbf{gap} in the relevant literature, since no existing review focuses specifically on this application and its particularities. Published surveys on DMs do not focus on image augmentation in a targeted manner, given the recent progress and achievements in the field. Existing surveys either focus on traditional image augmentation \cite{yang2022image}, or provide a general overview of DMs without delving into their specific application for image augmentation \cite{cao2024survey, yang2023diffusion, song2024lightweight}. For instance, the study of \cite{MUMUNI2022100258} presents a comprehensive taxonomy of image augmentation approaches, including input space transformations, feature space augmentation, data synthesis and meta-learning based methods. However, it does not cover the latest advancements concerning the use of DMs. In contrast, the work of \cite{croitoru2023diffusion} presents three generic diffusion modeling frameworks, which are based on denoising diffusion probabilistic models, noise conditioned score networks, and stochastic differential equations, but does not explore their potential for image augmentation. Some surveys have touched upon the efficiency aspect of DMs \cite{ulhaq2022efficient} or their application in specific domains like medical imaging \cite{kazerouni2023diffusion, kebaili2023deep}. However, these works do not provide a comprehensive overview of DMs for augmentation across various computer vision tasks. Furthermore, a recent survey categorizes augmentation methods based on large learning models, including those based on DMs, but does not focus specifically on them \cite{zhou2024survey}.

The above-described situation has \textbf{motivated} this study, in an attempt to remedy the identified gap. Its main \textbf{contribution} is a systematic, comprehensive and in-depth review of DM-based approaches specifically for image augmentation, covering a wide range of strategies, tasks and applications. In particular, the current work analyzes fundamental principles, introduces a detailed taxonomy of methods, examines evaluation metrics, discusses current challenges and provides perspectives on future research directions.

The remainder of this article is structured in the following way. Section \ref{sec:foundations} outlines the foundations and underlying principles of DMs. Section \ref{sec:categories} describes the different categories of DM-powered image augmentation methods, while Section \ref{sec:explain} presents and explains these categories in depth. Section \ref{sec:evaluation} surveys the various evaluation metrics used to assess the performance of DMs for image augmentation. Section \ref{sec:limitations} discusses the current challenges and limitations associated with the use of DMs for image augmentation. Finally, Section \ref{sec:conclusion} draws insights from the preceding discussion, along with suggestions for future research directions.

\section{Foundations of Diffusion Models}
\label{sec:foundations}

Diffusion Models (DMs) are a powerful class of generative models gaining significant traction in image synthesis. Inspired by non-equilibrium thermodynamics \cite{sohldickstein2015deep}, they operate by incrementally destroying structure in the data, through an iterative process of adding Gaussian noise (forward diffusion) that progressively transforms the data distribution towards a distribution of pure random noise. Then, a learnable reverse diffusion process that restores structure in the data yields a tractable generative model. Thus, DMs are trained for gradually transforming random noise patterns into samples of the data generating distribution. This section details the principles underlying DMs, clarifying why they are particularly suited for visual data augmentation.
\subsection{Forward Diffusion Process}
The Forward Diffusion (FD) process \cite{ho2020denoising} is the cornerstone of DMs, as it corrupts the training dataset by sequentially inserting Gaussian noise. Assume that the initial data distribution is $q(x_0)$, where subscript `$0$' denotes the original/unmodified state of the dataset and $x_0 \sim q(x_0)$ is an image from this dataset. FD proceeds as a sequence $q$ of incrementally noised versions, $x_1, x_2 \dots, x_T$, which are generated by a Markov chain. The conditional distribution for each step in this sequence, $p(x_t|x_{t-1})$, is modeled as a Gaussian $\mathcal{N}(x_t;\sqrt{1-\beta_t} x_{t-1}, \beta_t \mathbf{I})$, where $t$ ranges from $1$ to $T$ and it denotes the total noise added to the input image. $T$ corresponds to the total number of diffusion steps, $\beta_1,\dots,\beta_T$ is a sequence of variance parameters that define the noise level at each step, $\mathbf{I}$ is the identity matrix matching the dimensionality of input $x_0$ and $\mathbf{N}(x; \mu, \sigma)$ denotes the normal distribution with mean $\mu$ and covariance $\sigma$.

A key attribute of FD is that $x_t$ can be sampled at any arbitrary time step $t$ in closed form, using a reparameterization trick:

\begin{eqnarray}
\text{Let}\ a_t = 1 - \beta_t \text{,}~\bar{a}_t = \prod_{i=1}^t a_i \nonumber \\
\text{Then }
q(x_t|x_0) = \mathcal{N}(x_t|\sqrt{\bar{a}_t}x_0, (1-\bar{a}_t)\mathbf{I}) \nonumber \\
x_t = \sqrt{\bar{a}_t}x_0 + \sqrt{1-\bar{a}_t}\epsilon, \label{eq:one_step_sample}
\end{eqnarray}

\noindent where integer $t\in[1,N]$ and $\epsilon \sim \mathbf{N}(0,\mathbf{I})$. Thus, the noisy version $x_t$ can be directly obtained through a cumulative variance adjustment $\beta_t$, determined by sequence ${a_i}$, where $a_t = 1 - \beta_t$. This allows one to compute any noisy version $x_t$ from the original image $x_0$ in a single step, without having to iteratively generate the noisy version of all intervening time steps.

\subsection{Reverse Diffusion Process}
Following the corruption introduced by FD, the iterative Reverse Diffusion (RD) process aims at recovering the original dataset images from their noisy versions. Instead of directly generating images from noise patterns, a denoising learning model, which can be a DNN, iteratively predicts the noise pattern added to the data at each individual step of the FD process, starting from the final FD output, so that it can be removed. Progressive denoising gradually refines the image across $T$ consecutive steps. This is the so-called Denoising Diffusion Probabilistic Model (DDPM) formulation \cite{ho2020denoising}. Alternatively, the model can learn the so-called `score function', which is the gradient of the log probability density function of the data with respect to the input. Then, the model's predictions at each time step can be used to iteratively sample from the distribution, by following the gradient. Such a DM variant is called Score-based Generative Model (SGM) \cite{song2020score}.

The employed predictive DNN is usually a U-Net CNN \cite{ronneberger2015u}. Regarding the mathematical formulation, the RD process is defined as follows:
\begin{equation}
p_\theta(x_{0:T}) = p(x_T) \prod_{t=1}^T p_\theta(x_{t-1}|x_t),
\end{equation}
\noindent where $p_\theta(x_{t-1}|x_t) = \mathcal{N}(x_{t-1},\mu_\theta(x_t, t), \Sigma(x_t, t)$ \label{eq:model_sampling}.

In \cite{ho2020denoising}, the U-Net is trained by the following loss function:
\begin{equation}
L_{simple} = \mathbb{E}_{t,x_0, \epsilon}[||\epsilon - \epsilon_\theta(x_t, t)||^2_2]. \label{eq:L_simple}
\end{equation}
where $\epsilon$ represents the Gaussian noise added to image $x_0$ to obtain the noisy version $x_t$ and $\epsilon_\theta(x_t, t)$ denotes the noise predicted by the DNN parameterized by $\theta$, given the noisy image $x_t$ and the time step $t$. In the SGM formulation, $\epsilon_\theta(x_t, t)$ is the predicted score and, thus, after training, $\mu_\theta(x_t, t)$ can be approximated by a function of $\epsilon_\theta(x_t, t)$. Even though $L_{simple}$ (\ref{eq:L_simple}) does not offer a way to learn $\Sigma_\theta(x_t, t)$, it has been shown in \cite{ho2020denoising} that the best results are obtained by fixing the variance to $\sigma^2_t\mathbf{I}$, rather than learning it.

RD is an iterative process of $T$ consecutive time steps, starting from noise pattern $x_T$ and gradually recovering the original image $x_0$. At each time step $t$, $\mu$ and $\Sigma$ are computed and a new version of the output image is generated, which subsequently serves as input for the next time step $t$. This need for sequential generation across $T$ consecutive iterations is a significant limitation of DMs. One simple improvement is to reduce the number of sampling steps, from $T$ to $K$ evenly spaced real numbers between $1$ and $T$ \cite{nichol2021improved}. Alternatively, the non-Markovian Denoising Diffusion Implicit Models (DDIMs) \cite{song2022denoising} sample only across $S$ diffusion steps $[t_1, \dots, t_S] \subseteq [1, T]$ during generation:

\begin{equation}
\begin{aligned}
    x_{t-1} =\sqrt{\bar{a}_{t-1}} 
    \left(\frac{x_t - \sqrt{1-a_t} \epsilon_\theta(x_t)}{\sqrt{a_t}}\right) + \sqrt{1-a_{t-1}-\sigma^2_t} \epsilon_\theta(x_t) + \sigma_t \epsilon_t.
\end{aligned}
\end{equation}

\subsection{Guidance}
\textbf{Classifier Guidance} \cite{dhariwal2021diffusion} leverages a pretrained closed-set classifier to condition the RD process of a pretrained unconditional DM on a desired class label. The classifier model $p_\phi(y|x_t)$, where $\phi$ denotes its parameters, supports as many different class labels $y$ as the potential conditioning classes. With this approach and given the SGM formulation, the RD process is adjusted at each time step by the gradient of the log-probability $\nabla_{x_t}\log p_\phi(y|x_t)$ that steers sampling. Thus, $\mu_\theta(x_t, t)$ is approximated by:
\begin{equation}
\begin{aligned}
 \epsilon_\theta(x_t, t) + s * \nabla_{x_t}\log p_\phi(y|x_t),
\end{aligned}
\end{equation}

\noindent where $s$ is a scaling factor controlling the strength of guidance. This method ensures that the generated samples conform to the target class distribution, without any need to retrain the unconditionally trained DNN.

\textbf{Classifier-Free Guidance} \cite{ho2022classifier} eliminates the need for an explicit separate classifier model, by conditioning the DM on class labels directly during its training. The DM is trained with both conditional $\theta_c$ and unconditional $\theta_u$ objectives, alternating between conditioning on labels and generating without labels. At inference time, guidance is implemented by interpolating between the conditional and unconditional scores, so that $\mu_\theta(x_t, t)$ is approximated as follows:

\begin{eqnarray}
\nabla_{x_t} \log p_\theta(x_t|y) = \nabla_{x_t} \log p_\theta{}_c(x_t|y) + w (\nabla_{x_t} \log p_\theta{}_c(x_t|y) - \nabla_{x_t} \log p_\theta{}_u(x_t)),
\end{eqnarray}
where $w$ is a weight parameter that controls the strength of the guidance. 

\subsection{Diffusion Models in Latent Space}
Despite the faster RD process of DDIMs, image generation in pixel space and in an arbitrary resolution remains a significant bottleneck. To this end, Latent Diffusion Models (LDMs) \cite{rombach2022highresolution} have been introduced that operate in a latent space, in order to significantly accelerate the generation process. In particular, an LDM relies on an external autoencoder pretrained on a large-scale dataset. Its encoder $\mathcal{E}$ learns to map images $x \in D_x$ into a special latent code $z = \mathcal{E}(x)$ \cite{van2017neural, agustsson2017soft}. Its decoder $\mathcal{D}$ learns to map such low-dimensional latent representations back to pixel space, so that $\mathcal{D}(\mathcal{E}(x)) \approx x$. Thus, a regular DM or DDIM is trained to generate codes within the latent space. The resulting code can be mapped back to a realistic, high-dimensional image via the pretrained $\mathcal{D}$.

The LDM can be conditioned on class labels, segmentation masks, or even text, which guide the generation process. Let $c_\theta(y)$ be a model that maps a raw conditioning input $y$ to a conditioning vector\footnote{In the case of text prompts, this can be any text encoder.}. The LDM loss is then formulated as:
\begin{equation}
\label{eq:ldm_loss}
L_{LDM} = \mathbb{E}_{z \in \mathcal{E}(x),y,\epsilon\in\mathcal{N}(0,1),t}[||\epsilon - \epsilon_\theta(z_t, t, c_\theta(y))||_2^2],
\end{equation}
\noindent where $t$ is the time step, $z_t$ is the latent representation noised at step $t$, $\epsilon$ is the unscaled noise sample, and $\epsilon_\theta$ is the denoising network's prediction. Intuitively, the objective is to correctly remove the noise added to a latent representation of an image. During training, $c_\theta$ and $\epsilon_\theta$ are jointly optimized to minimize the LDM loss. At inference time, a random noise tensor is sampled and iteratively denoised to produce a new latent image $z_0$.

\subsection{Types of Conditioning}
DMs can be conditioned on various types of information to guide the generation process. The most common types of conditioning include:
\begin{itemize}
\item \textbf{Text Conditioning}: Models like Stable Diffusion \cite{rombach2022highresolution} use text prompts to guide image generation. The text is typically encoded using CLIP \cite{radford2021learning} or T5 \cite{raffel2020exploring} encoders, and the resulting embeddings condition the denoising process through cross-attention mechanisms.
\item \textbf{Class Label Conditioning}: Models can be conditioned on discrete class labels \cite{dhariwal2021diffusion}, enabling class-specific image generation. This is often implemented through class embeddings that are concatenated with or added to the model's intermediate features.
\item \textbf{Image Conditioning}: Reference images can guide the generation process through various mechanisms: direct concatenation with model features \cite{saharia2022palette}, CLIP image embeddings \cite{avrahami2023spatext}, or learned image encoders \cite{rombach2022highresolution}. This enables tasks like image-to-image translation and style transfer.
\item \textbf{Segmentation Map Conditioning}: Incorporating segmentation maps allows for control over the spatial layout and the structure of the generated images. Methods like \cite{wang2022pretraining} incorporate segmentation maps through specialized encoders of feature injection to control the spatial distribution of semantic classes. This enables precise control over object placement and scene composition.
\item \textbf{Pose and Structure Conditioning}: Human pose maps \cite{zhang2023adding}, edge maps, or semantic segmentation masks can guide the spatial structure of the generated images. These are typically processed through specialized encoders and injected into the model via cross-attention or feature concatenation.
\item \textbf{Multi-Modal Conditioning}: Certain models combine multiple mechanisms simultaneously. For example, combining text prompts with segmentation masks \cite{avrahami2023spatext} or reference images with class labels \cite{ruiz2023dreambooth} for more precise control over the generation process.
\end{itemize}
The most suitable type of conditioning depends on the specific application and the desired degree of control over the generation procedure. Recent work has focused on developing more sophisticated approaches to conditioning that enable finer-grained control, while maintaining generation quality \cite{zhang2023adding, zhao2024uni}.

\subsection{Diffusion Transformer}
The Diffusion Transformer (DiT) \cite{peebles2023scalable} architecture represents a significant departure from U-Net-based designs, replacing convolutional neural layers with Transformer blocks for processing visual data. DiT maintains the same high-level structure as previous DMs, but processes images as sequences of spatial patches in the vein of Vision Transformers (ViT) \cite{dosovitskiy2020image}. The architecture consists of: a) A patchification layer that splits images into non-overlapping patches and linearly projects them, b) A sequence of Transformer blocks with self-attention mechanisms, c) Time step and condition embeddings that are added to the patch embeddings, and d) A final decoder layer that reconstructs the spatial input image.
The key advantages of DiT over U-Net-LDM include better scaling properties, improved handling of long-range dependencies through self-attention, and more flexible integration of conditioning information \cite{hatamizadeh2025diffit, mo2023dit}. However, DiT typically requires more computational resources for training, due to the quadratic complexity of the self-attention operations. Recent work has shown that DiT-based models can achieve superior generation quality when scaled to sufficient size, leading to their wide adoption \cite{peebles2023scalable, fei2024scaling}.

\subsection{Foundation Diffusion Models for Image Generation}
% Conditional LDMs have boosted the development of the `Stable Diffusion' (SD) Foundation Model, a Text-to-Image generator (T2I) which was pretrained on the LAION-5B dataset \cite{schuhmann2021laion}. SD has dominated much of recent research related to generative image synthesis for image augmentation. Still, several attempts have been made to further improve it. For instance, `Stable Diffusion XL' (SDXL) \cite{podell2023sdxl} innovates over the basic SD in three ways:
% \begin{itemize}
%     \item It boasts a U-Net three times more complex and leverages a dual text encoding system for text conditioning. This new text encoder (OpenCLIP ViT-bigG/14), operating alongside the original one, significantly expands the model's capacity.
%     \item It enhances control over the final image crop, via the incorporation of size- and crop-conditioning during training. This is implemented by feeding crop parameters to the model as conditioning parameters via Fourier feature embeddings.
%     \item Its inference stage operates in two steps: a `base' model generates an initial image, which is then fed to a `refiner' model that adds finer, higher-quality details.
% \end{itemize}

\begin{sidewaystable*}[htbp!]
\centering
\caption{Overview of Foundation Diffusion Models (FDMs)}
\label{tab:foundation_models}
\tiny
\setlength{\tabcolsep}{5pt}
\begin{tabular}{p{1.6cm} p{0.8cm} p{0.8cm} p{2cm} p{2cm} p{2cm} p{0.8cm} p{0.9cm} p{3.3cm} p{2.5cm}}
\toprule
\textbf{Model} & \textbf{Release Date} & \textbf{License} & \textbf{Conditioning Mechanisms} & \textbf{Text Encoder} & \textbf{Training Data} & \textbf{Params} & \textbf{Res. (pixels)} & \textbf{Notable Features} & \textbf{Indicative Derived Methods} \\
\midrule \\ 
\multicolumn{10}{c}{\textbf{\footnotesize Pixel-Space U-Net FDMs}} \\ 
\hline
Imagen \cite{saharia2022photorealistic} & May'22 & Prop. & Text, Img & T5-XXL \cite{raffel2020exploring} & LAION-400M + Prop. ($\sim$ 860M) & 2B & $1024^2$ & First that explored using frozen LLMs as text encoders & \cite{wang2023imagen, chen2022re, hu2024instruct} \\
\midrule[0.03em]
DeepFloyd IF \cite{DeepFloydIF} & Apr'24 & Open & Text, Img & T5-XXL \cite{raffel2020exploring} & LAION-5B ($\sim$1B) \cite{schuhmann2022laion} & 400M - 4.3B & $1024^2$ & Cascaded pixel diffusion & \cite{tang2024iterinv}\\
\hline \\
\multicolumn{10}{c}{\textbf{\footnotesize Latent-Space U-Net FDMs}} \\
\hline
$SD 1.x$ \cite{rombach2022highresolution} & Dec'21 & Open & Text, Img, Mask, Layout, Semantic Map & $CLIP ViT-L/14$ \cite{radford2021learning} & LAION-5B \cite{schuhmann2022laion} & $860M$ & $512^2$ & First major open-source LDM & \cite{podell2023sdxl, cao2023masactrl, wei2023elite} \\
\midrule[0.03em]
$DALL-E 2$ \cite{ramesh2022hierarchical} & Apr'22 & Prop. & Text, CLIP Embeddings & $CLIP$ \cite{radford2021learning} & Prop. ($\sim 650$M) & $3.5B$ & $1024^2$ & Two-stage generation using CLIP latents & \cite{betker2023improving}\\
\midrule[0.03em]
$SD 2.x$ \cite{rombach2022highresolution} & Nov'22 & Open & Text, Img, Mask, Layout, Semantic Map & $OpenCLIP-ViT/H$ \cite{radford2021learning} & LAION-5B \cite{schuhmann2022laion} & 860M & $768^2$ & Improved text understanding & \cite{zhang2023adding, zhang2024forget} \\
\midrule[0.03em]
SD Cascade \cite{pernias2023wuerstchen} & Feb'23 & Open & Text, Img, Mask, Layout, Semantic Map, Edge Map & CLIP-H \cite{ilharco_gabriel_2021_5143773} & LAION-5B \cite{schuhmann2022laion} ($\sim$103M) & 4B & $1024^2$ & Highly compressed semantic latent space & \cite{phung2024coherent} \\
\midrule[0.03em]
SDXL \cite{podell2023sdxl} & Jul'23 & Open & Text, Img & CLIP ViT-L \cite{radford2021learning} + OpenCLIP ViT-bigG \cite{cherti2023reproducible} & Prop. & 2.6B & $1024^2$ & Dual text encoders with size and crop conditioning & \cite{sauer2025adversarial} \\
\midrule[0.03em]
Kandinsky 3 \cite{arkhipkin2023kandinsky} & Oct'23 & Open & Text, Img, Mask & Flan-UL2 \cite{tay2023new} & LAION-5B + COYO-700M + Prop. ($sim$150M) & 11.9B & $1024^2$ & Single-stage pipeline with large text encoder for better text understanding & \cite{patel2024lambda} \\
\midrule[0.03em]
$DALL-E 3$ \cite{betker2023improving} & Oct'23 & Prop. & Text, Img & $T5 XXL$ \cite{raffel2020exploring} & Prop. ($\sim 1$B) & - & $1024^2$ & Training on AI-generated detailed captions improves prompt following ability & - \\
\midrule[0.03em]
SDXL Turbo \cite{sauer2025adversarial} & Feb'24 & Open & Text, Img & CLIP-ViT-g-14 \cite{radford2021learning} & Distilling from SDXL & 3.5B & $512^2$ & Single-step generation & \cite{su2024generative} \\
\hline \\
\multicolumn{10}{c}{\textbf{\footnotesize Transformer-Based FDMs}} \\
\hline
PixArt-$\alpha$ \cite{chen2023pixart} & Sep'23 & Open & Text & Flan-T5-XXL \cite{raffel2020exploring} & SAM-LLAVA + JourneyDB + Prop. ($\sim$25M) & 600M & $1024^2$ & Efficient training with only 1\% cost of SOTA models & \cite{chen2024pixart} \\
\midrule[0.03em]
PixArt-$\delta$ \cite{chen2024pixart} & Jan'24 & Open &Text, Img & T5 \cite{raffel2020exploring} & Prop ($\sim$120K) & 600M & $1024^2$ & Fast 4-step sampling & - \\
\midrule[0.03em]
SD 3 \cite{esser2024scaling} & Mar'24 & Mixed & Text, Img & T5 XXL \cite{raffel2020exploring} & Prop. ($\sim 1$B) & $0.8B-8B$ & $1024^2$ & Rectified flow formulation enables efficient few-step sampling & - \\
\bottomrule
\end{tabular}
\end{sidewaystable*}

The evolution of DMs for image generation has been marked by several key architectural innovations and training strategies. These developments have resulted in the proliferation of commonly used Foundation Models (FMs) for image synthesis, which are available in a pretrained form and can be employed as a basis for more specialized methods. The architectures can be primarily categorized into \textbf{U-Net-based} \cite{rombach2022highresolution} and \textbf{Transformer-based} \cite{peebles2023scalable} ones, with \textbf{U-Net-based} methods further divided into those operating directly in \textit{pixel space} versus those performing inference in \textit{latent spaces}. All methods operate in the latent space of a pretrained autoencoder, unless those that perform directly in the pixel space. Moreover, U-Net-based methods employ a conditional LDM as backbone (or DDPM \cite{ho2020denoising}, if they operate in pixel space), while Transformer-based approaches utilize a conditional DiT as their backbone.\\

Table \ref{tab:foundation_models} illustrates key Foundation Diffusion Models (FDMs) presented in the literature, emphasizing on the following important characteristics: a) Release date, b) License, c) Conditioning mechanisms, d) Text encoder, e) Training dataset, f) Number of parameters, g) Resolution (in pixels), h) Notable features, and i) Indicative/important methods relying on them. As can be seen from Table \ref{tab:foundation_models}, most FDMs operate in latent space and use text/image conditioning, while U-Net is the dominant neural architecture before the recent emergence of Transformer-based approaches. LAION-5B \cite{schuhmann2022laion} and its variants serve as the predominant training dataset for many open-source FDMs, highlighting its significance in image generation. There is also an evolution in text encoders from CLIP-based models towards more sophisticated language models like T5-XXL, suggesting a trend toward better text understanding.

\begin{figure*}
    \centering
    \includegraphics[width=0.9\textwidth]{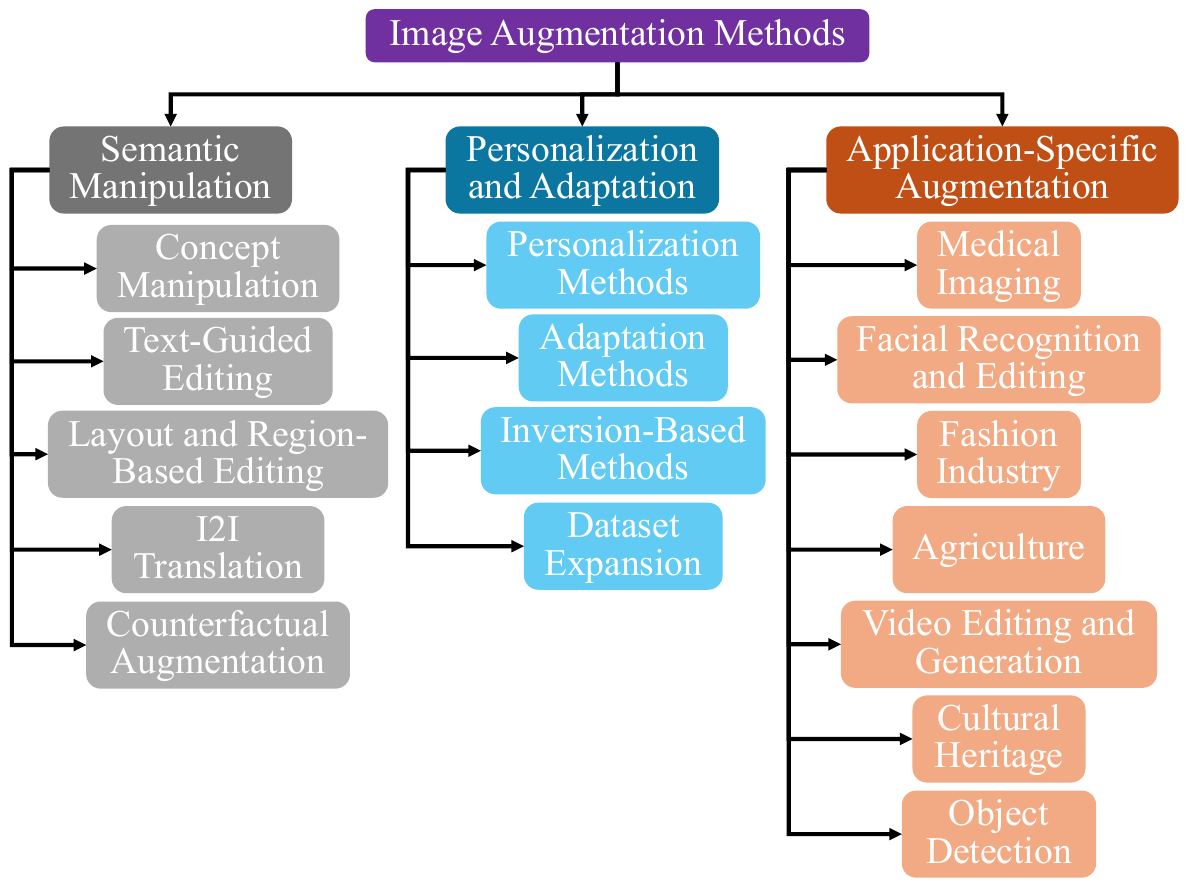}
    \caption{Taxonomy of DM-based image augmentation methods.}
    \label{fig:data_augmentation}
\end{figure*}

\section{Taxonomy of Diffusion Models for Image Augmentation}
\label{sec:categories}

\begin{figure*}
    \centering
    \adjustbox{center=\textwidth}{
        \includegraphics[width=1.05\textwidth]{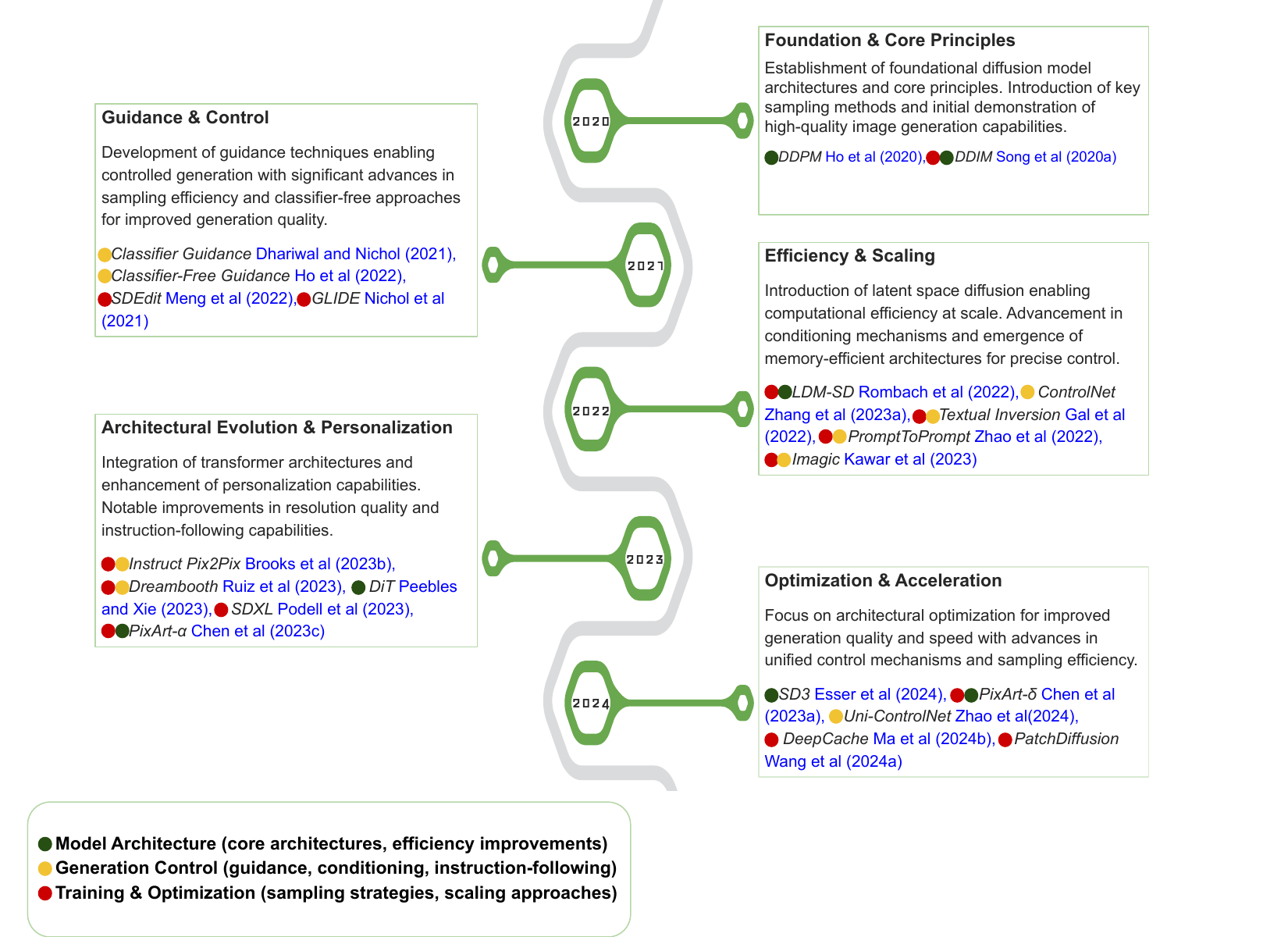}
    }
    \caption{Timeline representation of key recent DM-powered image augmentation methods}
    \label{fig:Timeline}
\end{figure*}

In this section, an overview of the landscape of DM-based methods for image augmentation is provided. In particular, a taxonomy of the various approaches is defined and graphically illustrated in Fig. \ref{fig:data_augmentation}. More specifically, considering as a criterion the task/goal of each method, DM-powered image augmentation approaches can initially be classified in the following main categories (while each class can be further divided into sub-categories, as will be discussed later in this section):
\begin{itemize}
\item \textbf{Semantic Manipulation}: The goal is to introduce fine-grained context-aware modifications to an image, while maintaining its main semantic contents \cite{kawar2023imagic, kim2022diffusionclip, zhang2023adding}.
\item \textbf{Personalization and Adaptation}: The target is to alter the appearance of the image, so as to better conform to specific datasets, tasks, requirements or user preferences \cite{ruiz2023dreambooth, gal2022image, wei2023elite}.
\item \textbf{Application-Specific Augmentation}: The goal is to regulate the augmentation process using domain specific knowledge, i.e. introducing modifications that are only meaningful for a given application (e.g., medical imaging, facial recognition, etc.) \cite{chambon2022roentgen, boutros2023idiffface}.
\end{itemize}

Complementarily to the categorization described above, Fig. \ref{fig:Timeline} illustrates a timeline representation that contains key recent DM-powered image augmentation methods. Each entry corresponds to a critical milestone work that significantly impacted the research field. Naturally, more recent works are associated with more complex and advanced DM models/architectures, leading also to superior performance. Moreover, Table \ref{tab:diffusion_augmentation} illustrates key representative methods, belonging to each category defined in the taxonomy of Fig.  \ref{fig:data_augmentation}.

\begin{table*}[htbp!]
\centering
\caption{DM-powered methods for image augmentation.}
\label{tab:diffusion_augmentation}
\footnotesize
\begin{tblr}{% TABLE SPECIFICATION
colspec={X[1.2,l,m] X[1.6,l,m] X[4.8,j,m]  }}
\hline     
\textbf{Category} & \textbf{Subcategory} & \textbf{Methods}\\\hline   
\SetCell[r=5]{}\begin{tabular}{c}Semantic\\Manipulation\end{tabular}
&\begin{tabular}{c}Concept\\Manipulation\end{tabular} & \begin{tiny}\cite{chen2024anydoor, luo2023camdiff, zhao2022x, zhang2024objectadd, song2022objectstitch, huang2023composer, brack2022stable,rando2022redteaming,schramowski2023safe,wasserman2024paint,gandikota2023erasing,gandikota2024unified,heng2024selective,kim2023safe,kumari2023ablating,ni2023degenerationtuning,zhang2024forget}\end{tiny}\\\hline 
&\begin{tabular}{c}Text-Guided\\Editing\end{tabular} & \begin{tiny}\cite{kawar2023imagic, yu2024uncovering, nichol2021glide, hertz2022prompt, chen2024training, lin2024text, huang2023region, brooks2023instructpix2pix, wang2023instructedit, yang2024editworld, jin2024reasonpix2pix, santos2024pix2pixonthefly, avrahami2023spatext, yang2023paint, kirstain2023xfuse, balaji2022ediff, geng2024instructdiffusion}\end{tiny}
\\\hline 
&\begin{tabular}{c}Layout and\\Region-Based\\Editing\end{tabular} & \begin{tiny}\cite{zeng2023scenecomposer, chen2023geodiffusion, xue2023freestyle, schnell2024scribblegen, lugmayr2022repaint, yu2023inpaint, ackermann2022highresolution, couairon2022diffedit, avrahami2022blended, avrahami2023blended, sarukkai2024collage, xie2023smartbrush, xiao2023fastcomposer, zhang2023adding,levin2023differential}\end{tiny} \\\hline 
&\begin{tabular}{c}I2I\\(Image-to-Image)\\Translation\end{tabular} & \begin{tiny}\cite{meng2021sdedit, xu2024cyclenet, kim2022diffusionclip, kwon2023diffusionbased, parmar2023zeroshot, su2022dual, tumanyan2023plug, wang2022pretraining, ma2023unified, trabucco2023effective, cao2023masactrl,saharia2022palette, Michaeli2024AdvancingFC, rahat2024data, lingenberg2024diagen} \end{tiny}\\\hline 
&\begin{tabular}{c}Counterfactual\\Augmentation\end{tabular} & \begin{tiny}\cite{sanchez2022healthy, sanchez2022diffusion,  madaan2023diffusion, yuan2022not, parihar2024balancing, vendrow2023dataset} \end{tiny}\\\hline 
\SetCell[r=4]{}\begin{tabular}{c}Personalization\\and\\Adaptation\end{tabular}
&\begin{tabular}{c}Personalization\\Methods\end{tabular} & \begin{tiny}\cite{ruiz2023dreambooth, ruiz2024hyperdreambooth, gal2022image, zhang2023prospect, kumari2023multi, vinker2023concept, sohn2023styledrop, dong2022dreamartist, chen2024subject, wei2023elite, gal2023encoderbased, shi2024instantbooth, tewel2023key, jia2023taming, chen2023disenbooth, han2023highly} \end{tiny}\\\hline 
&\begin{tabular}{c}Adaptation\\Methods\end{tabular} & \begin{tiny}\cite{hemati2023cross, wu2023uncovering, dunlap2022using, zang2023boosting, zhu2023domainstudio, qiu2023controlling, Islam2024DiffusemixLD, Islam2024GenMixED} \end{tiny}\\\hline 
&\begin{tabular}{c}Inversion-Based\\Methods\end{tabular} & \begin{tiny}\cite{zhou2023using, zhou2023training, zhang2023inversion, li2023stylediffusion, wallace2023edict, tang2024locinv, kwon2022diffusion, mokady2023null} \end{tiny}\\\hline 
&\begin{tabular}{c}Dataset\\Expansion\end{tabular} & \begin{tiny}\cite{zhang2211expanding, Li2023SemanticGuidedGI, ye2023synthetic, wang2022sindiffusion, bansal2023leaving, yin2023ttida, sheynin2022knndiffusion, blattmann2022semiparametric, Chen2024DecoupledDA} \end{tiny}\\\hline 
\SetCell[r=7]{}\begin{tabular}{c}Application-\\Specific\\ Augmentation\end{tabular}
&\begin{tabular}{c}Medical Imaging\end{tabular}& \begin{tiny}\cite{akrout2023diffusion, sagers2022improving, ali2022spot, pinaya2022brain, hu2022unsupervised, rouzrokh2022multitask, wolleb2022diffusion, chambon2022adapting, chambon2022roentgen, guo2023accelerating, xia2022low, packhauser2023generation} \end{tiny}\\\hline 
&\begin{tabular}{c}Facial Recognition\\and Editing\end{tabular} & \begin{tiny}\cite{boutros2023idiffface, huang2024data}\end{tiny}\\\hline 
&\begin{tabular}{c}Fashion Industry\end{tabular}& \begin{tiny}\cite{Li_2023_ICCV, kong2023leveraging}\end{tiny}\\\hline 
&\begin{tabular}{c}Agriculture\end{tabular}& \begin{tiny}\cite{deng2023stable, harnessing, chen2023deep}\end{tiny}\\\hline 
&\begin{tabular}{c}Video Editing\\and Generation\end{tabular} & \begin{tiny}\cite{shin2024edit, wu2023tuneavideo} \end{tiny}\\\hline 
&\begin{tabular}{c}Cultural Heritage\end{tabular}& \begin{tiny}\cite{cioni2023diffusion} \end{tiny}\\\hline 
&\begin{tabular}{c}Object Detection\end{tabular} & \begin{tiny}\cite{fang2024data, zhang2023diffusionengine, li2025simple, tang2024aerogen, ma2024erase}\end{tiny}\\\hline 
\end{tblr}               
\end{table*}

\subsection{Semantic Manipulation}
Methods belonging to this category aim to induce subtle and context-aware changes to an image, altering its interpretation or conveying additional contextual information, while preserving the core semantic content and maintaining a coherent and meaningful visual representation \cite{saharia2022palette, hertz2022prompt}. Such methods are useful for generating realistic and diverse training samples, while maintaining the original image context. More fined-grained and detailed sub-classes of this category are:

\begin{itemize}
\item \textbf{Concept Manipulation}: Concept manipulation involves altering the semantic content of an image, by adding, removing or modifying objects, attributes, or even the entire scene \cite{chen2024anydoor, song2022objectstitch}.
\item \textbf{Text-Guided Editing}: Text-guided editing leverages natural language descriptions to directly influence the editing process of images, allowing for precise control over the modifications based on textual inputs. This approach combines the strengths of Natural Language Processing (NLP) and computer vision technologies, enabling a nuanced interpretation of text into visual changes \cite{hertz2022prompt, brooks2023instructpix2pix, kawar2023imagic}.
\item \textbf{Layout and Region-Based Editing}: Layout and region-based editing involves modifying specific areas or rearranging elements within an image to alter its composition or focus. These methods are crucial for applications that require precise control over spatial arrangements and detailed modifications to image content \cite{avrahami2023blended, avrahami2023spatext, zeng2023scenecomposer}.
\item \textbf{I2I (Image-to-Image) Translation}: Image-to-Image (I2I) translation methods harness DMs to transform one source image into a different target one, maintaining the core content while altering its style, texture, or modal characteristics. This category is critical for applications ranging from artistic style transfer to functional medical imaging translations \cite{saharia2022palette, parmar2023zeroshot, trabucco2023effective}. 
\item \textbf{Counterfactual Augmentation}: Counterfactual augmentation uses DMs to generate images that represent hypothetical scenarios or what-if analyses, often used for enhancing model explainability and robustness. This includes generating counterfactual scenarios, where key elements are altered to assess potential outcomes. Such methods are useful in fields such as medical imaging and policy-making, where understanding the impact of variable changes is crucial \cite{sanchez2022diffusion, sanchez2022healthy}. 
\end{itemize}

\subsection{Personalization and Adaptation}
Personalization and adaptation methods tailor the augmentation process to better suit specific datasets, tasks or user preferences \cite{gal2022image, ruiz2023dreambooth}. These methods enhance the relevance and effectiveness of augmented data, by finetuning models to align with particular requirements.

\begin{itemize}
\item \textbf{Personalization Methods}: Personalization methods aim at adapting DMs to generate content that meets specific user needs or preferences. These approaches often involve finetuning models, leveraging textual or visual inputs, and optimizing for personalized outputs \cite{gal2022image, ruiz2023dreambooth, wei2023elite}. 
\item \textbf{Adaptation Methods}: Adaptation methods tailor DMs to different domains or tasks, improving their versatility and performance in new or varied contexts. These approaches are crucial for ensuring that models can generalize well across different datasets and applications \cite{qiu2023controlling, hemati2023cross}.
\item \textbf{Inversion-Based Methods}: Inversion-based DM methods leverage the ability to invert the diffusion process for enhancing image editing and augmentation capabilities, by obtaining the original image from a noisy or perturbed version \cite{mokady2023null, zhang2023inversion}.
\item \textbf{Dataset Expansion}: Dataset expansion methods utilize DMs to synthetically generate additional images, given an original input dataset. The goal is to address the limitations of small-scale datasets and to enhance the diversity of training images. This is critical for improving the generalization and robustness of machine learning models, particularly when the acquisition of extensive labeled datasets is impractical \cite{zhang2211expanding, Li2023SemanticGuidedGI}. 
\end{itemize}

\subsection{Application-Specific Augmentation}
Application-specific augmentation methods tailor the augmentation process so as to meet the unique requirements and properties of a given application domain, i.e. they extensively rely on using detailed and domain specific knowledge \cite{chambon2022roentgen, boutros2023idiffface}. Such particular characteristics can be exploited for improving or guiding the augmentation process. Typical domains with unique requirements or properties are medical imaging, facial recognition, fashion industry, agriculture, etc.

\begin{itemize}
\item \textbf{Medical Imaging}: DMs for image augmentation are extensively employed to generate high-fidelity synthetic medical images, enhance existing datasets, and improve the robustness of diagnostic models. These methods address challenges such as data scarcity, variability in medical conditions, as well as the need for anonymized training data \cite{chung2022mr, ali2022spot}.
\item \textbf{Other Domain-Specific Applications}: DMs have also been effectively applied to a wide range of other domain-specific applications, taking into account the specific requirements of each field. Examples include facial recognition and editing \cite{boutros2023idiffface}, fashion industry \cite{Li_2023_ICCV}, agriculture \cite{deng2023stable}, etc.
\end{itemize}

% {

\section{DM-Powered Methods for Image Augmentation}
\label{sec:explain}
This section details the fundamental principles and mechanisms of DM-based methods for image augmentation, based on the taxonomy of approaches discussed in Section \ref{sec:categories}. The main benefits of each (sub)category are highlighted; thus, providing key insights regarding their practical usage. 

\subsection{Semantic Manipulation}
Semantic manipulation transforms the image appearance, while partially preserving its semantic content \cite{brooks2023instructpix2pix}, or manipulates the depicted semantic concept, potential textual elements, or layout. As discussed in Section \ref{sec:categories}, its subcategories are concept manipulation, text-guided editing, layout and region-based editing, image-to-image (I2I) translation, and counterfactual augmentation. In many cases the goal is to transform a specific input `reference' image to an augmented variant, a process generally called `editing'.

\subsubsection{Concept Manipulation}
Several methods for concept manipulation specialize in placing objects within images using the pretrained SD model, often being themselves training-free \cite{chen2024anydoor,luo2023camdiff,zhao2022x,zhang2024objectadd}. In particular, they condition SD on new objects specified by text prompts, Web-retrieved images or a high-frequency map of the target object, stitched with the scene at the desired location (generated from Contrastive Language-Image Pre-training (CLIP) \cite{radford2021learning}) for placing them at the background of an input image. The new objects might be checked for semantic consistency via CLIP embedding similarity. For example, the method in \cite{chen2024anydoor} employs identity feature extraction (using a self-supervised DINOv2 model \cite{oquab2023dinov2}), detail feature extraction and feature injection to seamlessly integrate the target object into the scene, by feeding the ID tokens and the detail maps into the pretrained SD as guidance to generate the final composition. It also supports additional controls, like user-drawn masks to indicate the desired shape of the object during inference.

\begin{figure*}
    \centering
    \includegraphics[width=1.0\linewidth]{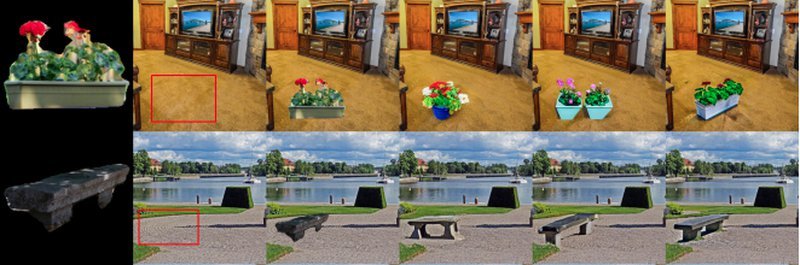}
    \caption{Comparison of various semantic manipulation methods: a) Desired object (col. 1), b) Target image and desired object location (col. 2), c) Copy-And-Paste (col. 3), d) BLIP \cite{li2022blip} (col. 4), e) SDEdit \cite{meng2021sdedit} (col. 5), f) ObjectStitch \cite{song2022objectstitch} (col. 6). Image from \cite{song2022objectstitch}.}
    \label{fig:ObjectStitch}
\end{figure*}

The method in \cite{song2022objectstitch} adapts the pretrained SD model for realistic object integration into scenes. It uses a content adaptor module that maps visual features from the input image object to a text embedding space to condition the SD. It is first pretrained on image-text pairs to learn semantics and then finetuned with SD to preserve object appearance. The SD model takes in the background image and the object embedding to generate a composite image. An example is depicted in Fig. \ref{fig:ObjectStitch}, where various methods are compared on object adding at a specific location in the target image. The first column displays the desired object, while the second one the target image and the desired location of the object.

The methods `Composer' \cite{huang2023composer} and `Stable Artist' \cite{brack2022stable} provide fine-grained control of the image generation process, by leveraging operations in the latent space. The `Composer' decomposes an image into representative factors, such as text description, depth map, sketch, color histogram, etc., and trains a DM (Guided Language to Image Diffusion for Generation and Editing (GLIDE) \cite{nichol2021glide}) conditioned on these factors, allowing customizable content creation by recombining the aforementioned factors. `Stable Artist' \cite{brack2022stable} uses Semantic Guidance (SEGA) to steer the diffusion process along multiple semantic directions corresponding to editing prompts, enabling subtle edits as well as changes in composition and style without masks or finetuning. SEGA allows the user to control the latent space representation, by calculating guidance vectors between the noise estimates of the original prompt and the editing prompts, and applying these to shift the unconditional noise estimate.

Large T2I models like SD can also replicate undesired behavior \cite{birhane2021large} or generate inappropriate content, such as copyrighted artworks \cite{10.1145/3600211.3604681} or explicit images \cite{schramowski2023safe}. To address these issues, several methods have been proposed, which can be organized in four main groups:\\

\noindent \textbf{Image Post-Processing} \cite{rando2022redteaming}: These methods filter out inappropriate content from the generated images after the generation process.\\
\textbf{Inference Guidance} \cite{schramowski2023safe}: These methods guide the diffusion process during inference to avoid generating undesired concepts. For example, Safe Latent Diffusion \cite{schramowski2023safe} defines an `unsafe' textual concept and uses it to guide the diffusion process away from generating inappropriate content.\\
\textbf{Image Inpainting} \cite{yu2023inpaint,wasserman2024paint}: These methods remove undesired objects or regions from images and inpaint the missing parts with appropriate content. They often use a mask to specify the region to be inpainted.\\
\textbf{Model Finetuning} \cite{gandikota2023erasing,gandikota2024unified,heng2024selective,kim2023safe,kumari2023ablating,ni2023degenerationtuning,zhang2024forget}: These methods finetune the pretrained DM to prevent it from generating undesired concepts. They often use approaches like concept erasure \cite{gandikota2023erasing}, self-distillation \cite{kim2023safe}, or degeneration-tuning \cite{ni2023degenerationtuning} to remove the unwanted concepts from the model's learned representations.

\subsubsection{Text-Guided Editing}
Unlike text-prompt-based concept manipulation, which directly uses the text prompt to guide the editing process, text-guided editing methods first optimize a text embedding to reconstruct the input reference image \cite{kawar2023imagic,yu2024uncovering,nichol2021glide}. The reconstruction is the output of a pretrained conditional LDM, i.e., the outcome of the RD process. Additionally, a target text embedding is the CLIP representation of a textual prompt, which is given as a condition to the LDM. In the end, these methods interpolate between the optimized embedding and the target text embedding, so that conditioning on this interpolated vector eventualy generates the desired edited/augmented image. For example, Imagic \cite{kawar2023imagic} optimizes a text embedding, finetunes the DM to better reconstruct the input image, and then linearly interpolates between the optimized embedding and the target text embedding. Subsequently, this interpolated embedding is provided as a condition to the finetuned DM, so that it generates the desired edited augmented image.

\begin{figure*}
    \centering
    \includegraphics[width=0.9\linewidth]{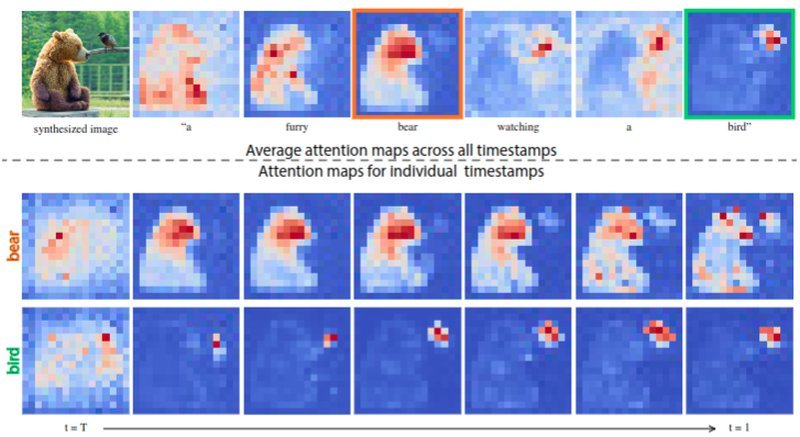}
    \caption{Cross-attention maps of a text-conditioned diffusion image generator. The top row illustrates the average attention masks for each word in the prompt that synthesize the image on the left. The bottom rows depict the attention maps from different diffusion steps with respect to the words `bear' and `bird'. Image from \cite{hertz2022prompt}.}
    \label{fig:prompttoprotmp}
\end{figure*}

A different approach to influence how the image is built word-by-word is by incorporating cross-attention maps within the diffusion process. These maps determine which parts of the image (`pixels') focus on specific elements (`tokens') of the text prompt at different stages of image creation, as illustrated in Fig. \ref{fig:prompttoprotmp}. Methods like \cite{hertz2022prompt} and \cite{chen2024training} use cross-attention maps to align textual descriptions with image regions, enabling nuanced image alterations.
Some methods focus on segmenting images into learnable regions that can be individually manipulated, based on text commands. For example, the methods of \cite{lin2024text} and \cite{huang2023region} divide the image into regions, by using pretrained models for feature extraction, and then apply text-guided editing to each region separately. This enhances the granularity of edits and allows for more precise control over specific parts of the image.

In general, editing images based on human instructions allows a higher level of control over the DM's actions. These methods typically utilize a Large Language Model (LLM) to guide the editing process via human-like instructions \cite{brooks2023instructpix2pix,wang2023instructedit,yang2024editworld,jin2024reasonpix2pix,santos2024pix2pixonthefly,geng2024instructdiffusion}. For example, InstructPix2Pix \cite{brooks2023instructpix2pix} makes use of GPT-3 to generate a synthetic dataset of editing instructions (e.g., input caption: `photograph of a girl riding a horse', edited caption: `photograph of a girl riding a dragon') and subsequently a pretrained SD model is utilized to generate synthetic images matching the captions before/after. Subsequently, an SD model is trained on this generated dataset (by optimizing the LDM loss), so as to learn to perform image edits. Classifier-free guidance can also be used along with a text prompt to guide the diffusion process and to generate new images that comply better with their conditioning \cite{bansal2023universal}. This improves the alignment between the generated images and the text prompt, without requiring a separate classifier model.

Moving towards a different direction, the method of \cite{avrahami2023spatext} introduces a `spatio-textual representation' that combines CLIP image embeddings of object segments during training and CLIP textual embeddings of local prompts during inference. This representation is incorporated into the DM, by concatenating it with the noisy image or latent code, and the DM is finetuned accordingly (by minimizing the LDM loss). Multi-conditional classifier-free guidance is employed to control the relative importance of each condition, involving a fine-grained variant using separate guidance scales and a fast variant using a single scale for the joint probability of all conditions.

Paint-by-Example \cite{yang2023paint} introduces an exemplar-based editing approach, where it automatically merges a reference image into a source one. Self-supervised training utilizes the object's bounding box as a binary mask and the image patch inside as the reference image to reconstruct the source image. An information bottleneck compresses the reference image with a CLIP class token and injects the decoded feature into the diffusion process for better content understanding. Strong augmentations reduce train-test mismatches and handle irregular mask shapes. Controllability is achieved by training with irregular masks and using a classifier-free sampling strategy to adjust similarity between the edited and reference images.

X\&Fuse \cite{kirstain2023xfuse} presents a general approach for conditioning on visual information in T2I generation: processing a reference and the input image separately, using shared ResBlocks, and then concatenating them (before attention blocks) to allow interaction between the two images. eDiff-I \cite{balaji2022ediff} introduces an ensemble of expert denoisers, each specialized for a particular stage of the generation process, to capture distinct behaviors and to enhance model capacity without increasing computational cost during inference. The study compares T5 textual embeddings \cite{raffel2020exploring}, CLIP textual embeddings and optionally CLIP image embeddings during training. Subsequently, an inference approach (Paint-with-words) is presented, where the user can select phrases from the text prompt and can create binary masks of objects, which are provided as inputs to the model in order to control the spatial location of the objects. This is performed by modifying the cross-attention matrix between the image and the text features.

\subsubsection{Layout and Region-Based Editing}
\begin{figure*}
    \centering
    \includegraphics[width=1.0\linewidth]{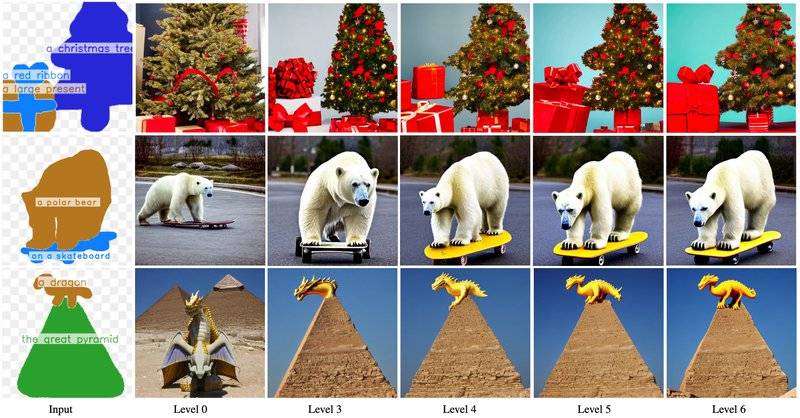}
    \caption{Indicative layout and region-based editing results with different precision levels. For each input layout, the images are progressively modified (starting with same noise), so that the generated images at different precision levels can have similar styles. Image from \cite{zeng2023scenecomposer}.}
    \label{fig:SceneComposer}
\end{figure*}

Several methods use text prompts to guide the generation and manipulation of images, based on layout and region information. For example, the approaches of \cite{zeng2023scenecomposer} and \cite{chen2023geodiffusion} enable semantic image synthesis and geometric control using text prompts. These methods often utilize a separate layout encoder to model spatial and semantic information into a format suitable for image generation. Indicatively, the latter consists of a precision-based mask pyramid that represents region shapes at multiple resolutions combined with text embeddings in \cite{zeng2023scenecomposer}. An example of such an approach is depicted in Fig. \ref{fig:SceneComposer}. In \cite{chen2023geodiffusion}, the layout encoder translates geometric layouts into text prompts by mapping locations, classes, and conditions into text tokens, in order to extract spatial information from the input image's layout, and then conditions the DM on this layout encoding.

ControlNet \cite{zhang2023adding} is another powerful approach for layout and region-based editing. It introduces a DNN architecture that adds spatial conditioning controls to large, pretrained T2I DMs, like SD. ControlNet creates a trainable copy of the model's encoding layers and connects it to the original model using `zero convolutions'. This allows for efficient finetuning on small datasets for various conditioning tasks, such as edge detection, pose estimation, and depth mapping. The architecture can process both text prompts and conditioning images (e.g., edge maps, pose maps, depth maps) as inputs, making it highly versatile for different types of spatial control in image generation and editing.

Certain methods focus on generating images from coarse layouts or scribbles. For instance, the method of \cite{xue2023freestyle} utilizes SD to generate freestyle images, by integrating semantic text embeddings with spatial layouts into SD. It represents each semantic class in the input layout using a text concept, which are then encoded into text embeddings. A Rectified Cross-Attention (RCA) module is subsequently introduced to inject these text semantics into the corresponding layout regions within the diffusion model's U-Net cross-attention layers. By finetuning just the U-Net with the integrated RCA on layout-image pairs, the pretrained model can generate images conditioned on both user-specified layouts and free-form text prompts, enabling capabilities like binding new attributes to objects and generating unseen object classes. The method of \cite{schnell2024scribblegen} utilizes ControlNet \cite{zhang2023adding} to generate synthetic images conditioned on scribble labels and text prompts. It employs classifier-free guidance (10\% of the conditioning scribble inputs are randomly dropped and replaced with a learnable embedding) and introduces an encoding ratio to adjust the diversity and photorealism of the generated images (by performing fewer FD steps), allowing for a trade-off between mode coverage and sample fidelity.

Inpainting methods \cite{lugmayr2022repaint,yu2023inpaint} make use of DMs to fill-in missing regions of an image based on a given mask. They typically condition the DM on the masked image and the mask itself, and then generate content to fill-in the masked region. A subset of methods \cite{ackermann2022highresolution,couairon2022diffedit,avrahami2022blended,avrahami2023blended} employs a multi-stage diffusion process with mask guidance, in order to achieve high-resolution image editing and inpainting.

Collage Diffusion \cite{sarukkai2024collage} is a method that takes a user-defined sequence of layers as input, called Collage. It consists of a full-image text string describing the entire image to be generated, along with a sequence of layers ordered from back to front; each layer consists of an RGBA image (alpha-masked input image) and a text describing it. The method modifies the text-image cross-attention in the DM to achieve spatial fidelity, while extending ControlNet to preserve appearance fidelity on a per-layer basis. Collage Diffusion also allows users to control the harmonization-fidelity trade-off for each layer by specifying desired noise levels, enabling layer-by-layer image editing. Similarly, SmartBrush \cite{xie2023smartbrush} utilizes text and shape guidance for object inpainting with DMs, enabling users to control the inpainted content based on both text descriptions and object masks. FastComposer \cite{xiao2023fastcomposer} is a tuning-free multi-subject image generation method that augments text prompts with visual features extracted from input  images using an image encoder. It models text prompts and input images as embeddings, using pretrained CLIP encoders, and then uses an Multilayer Perceptron (MLP) to augment the text embeddings with the visual features. The method trains the image encoder, MLP module, and U-Net with a denoising loss using a subject-augmented image-text paired dataset, while localizing cross-attention maps using the reference subject's segmentation mask to prevent identity blending in multi-subject generation. It also employs delayed subject conditioning in iterative denoising to balance identity preservation and editability.

The method of \cite{levin2023differential} is designed for allowing editing with pixel-level control over the amount of applied modification, using a `change map'. The latter is a matrix of dimensions identical to the spatial resolution of the original input image, describing the strength of the edit to be applied at each location. Different approaches can be used to generate it, such as Segment-Anything \cite{kirillov2023segment}, MiDas \cite{ranftl2020robust} or even manually drawn change maps. The method can operate in latent space and has been evaluated on SD, SDXL, Kandinsky \cite{arkhipkin2023kandinsky} and DeepFloyd IF \cite{DeepFloydIF} pretrained LDMs, using text prompts (either manually created, or by reversing the input image into CLIP and BLIP \cite{li2022blip}) along with change maps to guide the inference process.

\subsubsection{I2I (Image-to-Image) Translation}
\begin{figure}
    \centering
    \includegraphics[width=0.6\linewidth]{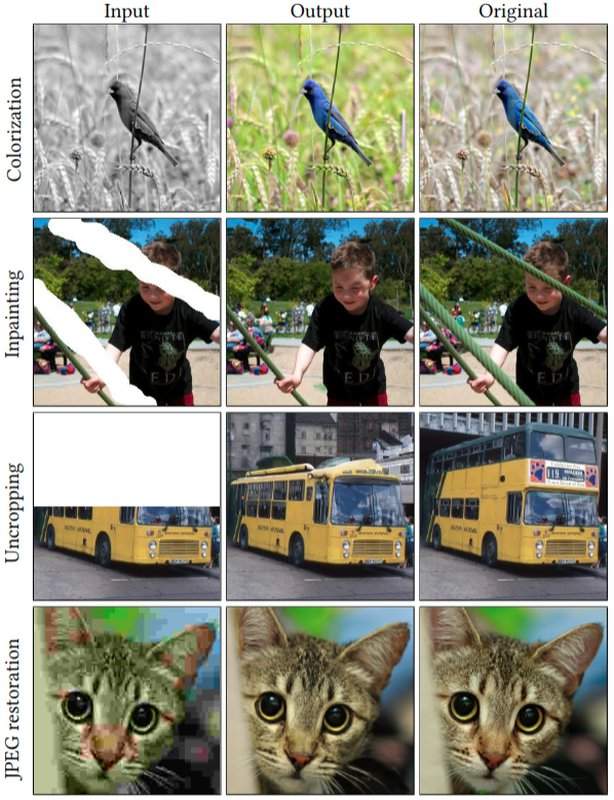}
    \caption{Indicative examples of an Image-to-Image DM that can generate high-fidelity outputs across a variety of tasks, without incorporating task-specific customizations. Image from \cite{saharia2022palette}.}
    \label{fig:I2IExample}
\end{figure}

Several methods make use of conditional DMs for I2I translation \cite{Michaeli2024AdvancingFC, rahat2024data}. For example, SDEdit \cite{meng2021sdedit} utilizes a conditional DM, guided by stochastic differential equations, to perform image editing. CycleNet \cite{xu2024cyclenet} and DiffusionCLIP \cite{kim2022diffusionclip} finetune a conditional DM with a CLIP-based loss to ensure that the generated image matches the target image's text description. Palette et al. \cite{saharia2022palette} employ conditional DMs to learn the distribution $p(y|x)$ for various tasks, like colorization, inpainting, JPEG restoration, and uncropping. The model incorporates a U-Net architecture with self-attention layers, conditioning on the input image through concatenation. During training, it predicts the noise added to the original image, minimizing L2 or L1 loss between the predicted and the actual noise. Then, the inference process involves iterative denoising over 1000 timesteps, starting from Gaussian noise and progressively refining the output. The method's strength lies in its ability to handle multiple tasks with a single architecture, eliminating the need for task-specific customizations (Fig. \ref{fig:I2IExample}).

A different group of methods focus on disentangling style and content for more targeted I2I translation \cite{kwon2023diffusionbased, lingenberg2024diagen}. For instance, the approach of \cite{kwon2023diffusionbased} makes use of separate encoders for style and content, and then injects these representations into the DM to generate the translated image. Other methods explore few-shot or zero-shot I2I translation. For example, pix2pix-zero \cite{parmar2023zeroshot} first inverts the input image to obtain a noise map, using DDIM and BLIP. These noise maps are then regularized to improve editability, using an autocorrelation objective. Edit directions are automatically discovered by generating diverse sentences for the source and target domains, and computing the mean difference between their CLIP embeddings. In order to preserve image structure during editing, a novel cross-attention guidance technique is employed, which matches the edited cross-attention maps with reference maps corresponding to the original structure. The method builds upon the SD model, utilizing its CLIP conditioning and cross-attention layers. On the other hand, Dual Diffusion Implicit Bridges (DDIBs) \cite{su2022dual} encompass two independently trained DMs (one for the source domain and one for the target one) to perform I2I translation, without requiring joint training on paired data.

The method of \cite{tumanyan2023plug} injects spatial features and self-attention maps extracted from the diffusion process of a source image into the generation process of the translated image, enabling fine-grained control over the generated structure, while complying with the target text prompt. Additionally, the method of \cite{wang2022pretraining} leverages a pretrained T2I DM (GLIDE) as a generative prior, consisting of a base model at $64$x$64$ resolution and an upsampling model to $256$x$256$. The framework adopts an encoder-decoder architecture, where a task-specific encoder maps input conditions (e.g., segmentation masks, sketches) to the semantic latent space of the pretrained diffusion decoder. The DM is pretrained on diverse text-image pairs, enabling the latent space to be conditioned on highly semantic text embeddings. The pipeline involves pretraining the DM, training a task-specific encoder using a two-stage finetuning scheme, finetuning the diffusion upsampler with adversarial training, and sampling the base model and upsampler during inference, using the encoded input and normalized classifier-free guidance. Moreover, another relevant framework is presented in \cite{ma2023unified}, aiming at joint subject- and text-conditional image generation. The DM (SD) generates high-quality images that align semantically with input texts, while preserving input image subjects. It leverages CLIP encoders to map texts and images into a unified multi-modal latent space. A fusing sampling strategy balances noise predictions between the unified condition and pure text condition to avoid overfitting to background information, i.e., the final noise estimate fuses the two predictions using a given ratio.

DA-Fusion \cite{trabucco2023effective} is another method that aims to to alter the semantic contents of an image. It relies on the pretrained SD and modifies the input reference image based on class label guidance. Novel visual concepts outside the DM's training dataset are handled by inserting and finetuning new token embeddings in the text encoder, using the textual inversion approach (see Section \ref{ssec::PersonalizationAdaptation}). Input images are spliced into the diffusion process at a random timestep to guide the generation, rather than generating from scratch. Randomizing the splice timestep provides diverse augmentation intensities.

MasaCtrl \cite{cao2023masactrl} introduces a mutual self-attention mechanism within U-Net, allowing the model to query correlated local structures and textures from a source image, while maintaining consistency with a target editing prompt. The architecture modifies the standard U-Net blocks by converting self-attention into mutual self-attention, where query features originate from the current denoising process, while key and value features are derived from the input image's diffusion process. In order to prevent confusion between foreground and background elements, MasaCtrl employs a mask-guided mutual self-attention strategy, utilizing masks extracted from cross-attention maps.

\subsubsection{Counterfactual Augmentation}
\begin{figure*}
    \centering
    \includegraphics[width=0.9\linewidth]{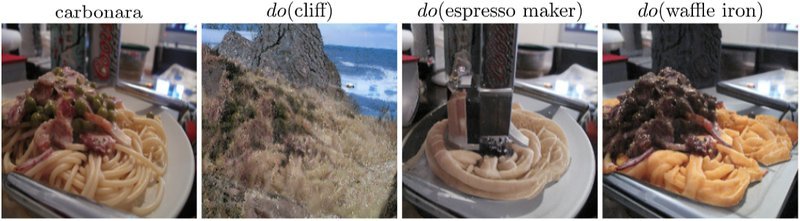}
    \caption{Counterfactuals on ImageNet generated by Diff-SCM. From left to right: a random image sampled from the data distribution and its counterfactuals do(class), corresponding to `how the image should change in order to be classified as another class?'. Image from \cite{sanchez2022diffusion}.}
    \label{fig:CounterfactualExamples}
\end{figure*}

A significant number of methods make use of conditional DMs to generate counterfactual images. For instance, the approach of \cite{sanchez2022healthy} utilizes a conditional DM to generate `healthy' counterfactuals of brain MRI images with tumors, by conditioning the model on the tumor mask. Diff-SCM \cite{sanchez2022diffusion} employs a conditional DM to estimate the effects of interventions in a causal framework, by conditioning the model on the intervention variable $do(class)$, corresponding to `how the image should change in order to be classified as another class'. For instance, for a given photo of a dog in a park, the dog should be replaced by a cat, while leaving the rest of the image unchanged if $do(cat)$ holds (Fig. \ref{fig:CounterfactualExamples}).

A great portion of methods focus on generating counterfactuals for specific applications. For example, %MEDJOURNEY \cite{gu2023medjourney} generates counterfactual medical images by conditioning the DM on a text description of the desired change, while
the methods of \cite{madaan2023diffusion} and \cite{yuan2022not} generate counterfactual images to explain the model's behavior and to improve its robustness. On the other hand, a different set of approaches utilize counterfactual augmentation to address bias and fairness issues in datasets and models. For example, the work of \cite{parihar2024balancing} utilizes a conditional DM to generate counterfactual images that balance the distribution of sensitive attributes (e.g., gender, race) in the dataset. The model is conditioned on the desired attribute distribution and generates images that match it, while preserving the remaining aspects of the image.

Another set of methods focus on expanding a given dataset with generated counterfactual or Out-of-Distribution (OOD) examples, in order to improve the robustness of a model (e.g., classifier) trained on the expanded dataset. For example, the method of \cite{vendrow2023dataset} utilizes a conditional DM to generate counterfactual examples that change specific attributes of the input image (e.g., object position, color, texture), while preserving the overall scene structure. These counterfactual examples are used to diagnose and mitigate model failures on OOD data.

\subsection{Personalization and Adaptation}
\label{ssec::PersonalizationAdaptation}
Personalization and adaptation constitute very common methodologies in image augmentation tasks. In the remainder of this subsection, respective DM-based approaches are detailed, grouped according to the taxonomy presented in Section \ref{sec:categories}.

\subsubsection{Personalization Methods} \label{sec:personalization}
Personalization implies that the DM must be adapted to generate content that meets specific user needs or preferences. These methods often involve finetuning pretrained models, leveraging textual or visual inputs, and optimizing for personalized outputs.

\begin{figure*}
    \centering
    \includegraphics[width=1.0\linewidth]{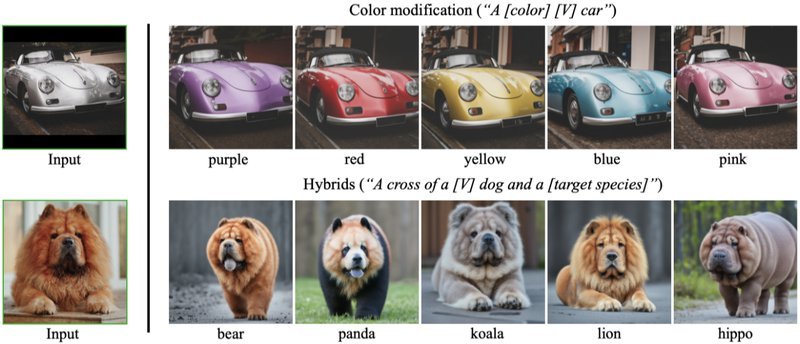}
    \caption{Example of personalization-oriented image augmentation, where the given subject's properties need to be modified, while preserving its key visual features that forge its identity. Image from \cite{ruiz2023dreambooth}.}
    \label{fig:PersonalizationDreambooth}
\end{figure*}

A significant number of methods make use of finetuning to personalize pretrained T2I DMs for subject-driven generation. For example, DreamBooth \cite{ruiz2023dreambooth} finetunes the DM within a few-shot learning setting, using a small set (3-5) of images of a single subject paired with a text prompt containing a unique identifier and the name of the class the subject belongs to (e.g., `A [V] dog'). These images contain a specific subject and serve as the training dataset that the DM needs to learn, while the text prompt simply describes the new subject. In particular, the method applies a class preservation loss to ensure that the generated images maintain the identity of the subject. Similarly, HyperDreamBooth \cite{ruiz2024hyperdreambooth} is a method for efficient personalized DMs that introduces three key components, namely Lightweight DreamBooth (LiDB), a HyperNetwork for fast personalization, and rank-relaxed fast finetuning. LiDB decomposes the rank-1 LoRA \cite{hu2021lora} weight space using a random orthogonal incomplete basis, resulting in smaller personalized models. Additionally, the HyperNetwork, consisting of a ViT encoder and transformer decoder, predicts LiDB residuals from input images, using diffusion denoising and weight-space losses. Eventually, rank-relaxed fast finetuning captures fine-level details by relaxing the LoRA rank and finetuning with the predicted HyperNetwork weights, improving subject fidelity while maintaining fast personalization (Fig. \ref{fig:PersonalizationDreambooth}).

A different set of methods make use of the so-called `textual inversion' \cite{gal2022image} to personalize T2I generation. This process learns a new text embedding (termed `pseudo-word') that represents a specific visual concept, using a small set of images depicting that concept. This allows the model to generate images of the concept, using natural language descriptions that include the pseudo-word. ProSpect \cite{zhang2023prospect} extends this idea by learning a collection of pseudo-words (called a `prompt spectrum') that capture different visual attributes (e.g., material, style, layout) of the concept.

On a different basis, another group of methods focus on combining multiple concepts for personalized generation. For example, the approaches of \cite{kumari2023multi} and \cite{vinker2023concept} utilize a compositional mechanism, where different concepts are represented by separate text embeddings that can be combined to generate novel images. StyleDrop \cite{sohn2023styledrop} and DreamArtist \cite{dong2022dreamartist} adopt a style-based approach, where the style of the generated image is controlled by a learned embedding.

Another conceptualization comprises methods that make use of encoder-based approaches to personalize T2I generation. For example, ELITE \cite{wei2023elite} utilizes an image encoder to map user-provided input images to text embeddings, which are then used to condition the pretrained SD. This allows the model to generate personalized images that match the style and content of the input. Similarly, the methods of \cite{gal2023encoderbased} employs a cross-attention mechanism to inject visual features from an image encoder into the DM at different layers.

% Several recent methods have further advanced personalized T2I generation.
SuTI \cite{chen2024subject} is an apprenticeship learning framework that trains expert DMs (Imagen \cite{saharia2022photorealistic}) for each subject on image-text clusters, where each cluster contains 3-10 images-prompts. Then, $K$ expert DMs are fine tuned on $K$ clusters in order to learn particular subjects belonging to the clusters. Subsequently, a single DM is trained on the outputs of these expert DMs. This approach enables SuTI to compose specialized knowledge from numerous expert models into a single, versatile model. During inference, it can generate customized images for new subjects, without some kind of optimization.

InstantBooth \cite{shi2024instantbooth} is an alternative approach that enables personalized generation without requiring deployment-time finetuning of the DM itself. It introduces a learnable image encoder to convert input images to a textual embedding and adapter layers to inject rich visual features for better identity preservation. The original model weights are frozen and only the new components are trained. A unique identifier token $V$ represents the input concept in the text prompt, while the input images are cropped and background-masked.

Perfusion \cite{tewel2023key} is a compact and efficient architecture based on key-locked rank-1 editing, which addresses key challenges in personalized generation, such as overfitting and maintaining high visual fidelity, while allowing creative control. Key-locked rank-1 editing prevents overfitting by locking the keys of a learned concept to those of its supercategory in the cross-attention layers, ensuring the concept inherits the generative prior without deviating too much. The method maintains high visual fidelity, by learning concept-specific values in an extended latent space to capture the concept's unique appearance. Moreover, the rank-1 update is gated based on the similarity between the current encoding and the target concept, allowing fine-grained control over the influence of each concept at inference time. 

Taming Encoder \cite{jia2023taming} is a method for generating images of customized objects specified by users, without the need for lengthy per-object optimization, as required by previous approaches. In particular, given an input image and a text description of the desired output, the image object encoder (CLIP) computes the object embedding, while the text encoder computes the text embedding. These two are then passed to Imagen \cite{saharia2022photorealistic} for generating the final output in a single forward pass. Training data is prepared using a captioning model (PaLI \cite{chen2022pali}) and binary masks to isolate the object. The framework is jointly trained on domain-specific and general-domain datasets, using cross-reference regularization and object-embedding dropping. The entire network, except from the object encoder, is tuned using the same objective as the pretrained DM.

DisenBooth \cite{chen2023disenbooth} is a framework for subject-driven T2I generation that utilizes a textual identity-preserving embedding and a visual identity-irrelevant embedding. In particular, an identity-preserving branch maps the subject's identity to a special text token using a CLIP text encoder, while an identity-irrelevant branch employs a pretrained CLIP image encoder and a learnable mask to extract identity-irrelevant features, which are then aligned with the text feature space using an MLP with skip connection (Adapter). During finetuning, DisenBooth utilizes denoising, weak denoising, and contrastive embedding objectives, and employs parameter-efficient finetuning using LoRA. After finetuning, DisenBooth enables flexible and controllable generation, by combining the subject's identity token with other text descriptions and optionally inheriting characteristics from a reference image. 

HiPer \cite{han2023highly} personalizes pretrained SD models for text-driven image manipulation, using a single input image. It decomposes the CLIP text embedding space into initial tokens and end tokens for preserving subject identity. In particular, given an input image, the input image's prompt, and the target desired image's prompt, HiPer optimizes the embedding in the DM's latent space. During inference, the target prompt's embedding is concatenated with the optimized HiPer embedding to condition the pretrained SD model, generating a manipulated image that preserves the subject's identity, while incorporating the target prompt's semantics. HiPer's main contribution lies on its decomposition of the text embedding space for separate control of semantics and identity, requiring only a single source image without finetuning the DM.

\subsubsection{Adaptation Methods}
\begin{figure*}
    \centering
    \includegraphics[width=1.0\linewidth]{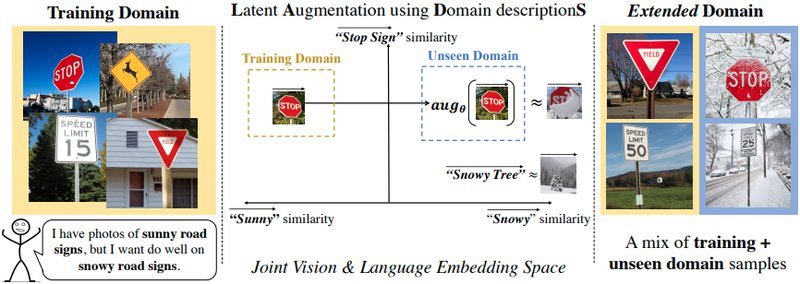}
    \caption{Example of DM-based domain adaptation for image augmentation, where a model trained on images from a source domain can be applied to images from a target one. Image from \cite{dunlap2022using}.}
    \label{fig:Adaptation}
\end{figure*}

A significant portion of adaptation methods rely on finetuning to adjust pretrained DMs to new domains. For example, the method of \cite{hemati2023cross} employs the pretrained SD model to generate synthetic images that bridge gaps between domains and reduce non-iidness in training data. It receives an image as input from one domain and a guidance attribute (either a text prompt or an image) from another one of the same class, using the LDM to create interpolated synthetic images. The approach of \cite{wu2023uncovering} introduces an optimization method to find the optimal soft combination weights of the input image description and the target image description text embeddings at each denoising step. The weights are optimized to match the target attribute using a CLIP-based loss, while preserving other content using a perceptual loss. The method of \cite{dunlap2022using} makes use of a domain-specific CLIP model to guide the diffusion process towards generating images that match the target domain (Fig. \ref{fig:Adaptation}). Moreover, the method of \cite{zang2023boosting} utilizes a domain-specific discriminator to filter out low-quality or out-of-domain samples during training.

Another group of approaches focuses on adapting pretrained DMs to specific tasks or applications. Indicative examples are the cases of nuclei segmentation \cite{yu2023diffusion} or knee osteoarthritis severity classification \cite{chowdary2023enhancing}, by finetuning on datasets of nuclei images or X-ray images, respectively. 

DomainStudio \cite{zhu2023domainstudio} comprises a recent method that adapts pretrained DMs to target domains using limited data, by introducing a pairwise similarity loss to maintain the relative distances between the generated samples during domain adaptation and designing a high-frequency detail enhancement approach to improve generation quality. Additionally, Orthogonal Finetuning (OFT) \cite{qiu2023controlling}, which is a method for adapting DMs to new tasks, injects trainable orthogonal matrices into the model's attention layers, transforming neuron weights multiplicatively. These matrices, initialized as identity and kept orthogonal, enable stable finetuning that preserves the model's semantic knowledge. OFT receives subject images or control signals with text prompts as input and generates images following the subject identity or control while matching the prompt. The orthogonal transformation of neurons maintains their pairwise angles, leading to faster convergence, better quality and controllability.

The methods of \cite{Islam2024DiffusemixLD} and \cite{Islam2024GenMixED} introduce a prompt-guided approach that enhances both in- and cross-domain image classification. Specifically, they leverage image editing with carefully designed conditional prompts to generate augmented images based on concatenation of original and edited content, followed by blending with fractal patterns, which reduces overfitting and ensures robust training.

\subsubsection{Inversion-Based Methods}
\begin{figure*}
    \centering
    \includegraphics[width=1.0\linewidth]{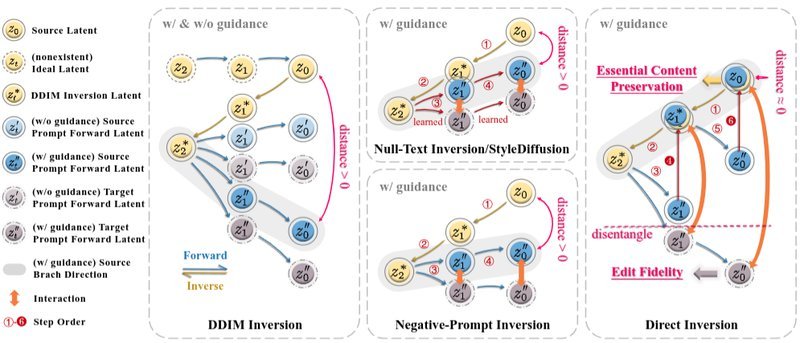}
    \caption{Various types of inversion-based approaches for diffusion-powered image editing. Image from \cite{ju2023direct}.}
    \label{fig:InversionMethods}
\end{figure*}

Inversion-based methods typically optimize a latent code, i.e., the latent-space representation of a user-provided input image that contains an object to be learnt by the DM, so that it can reconstruct the image using the DM. Subsequently, the latent code is manipulated so as to steer the DM to perform the desired editing or augmentation. Various types of inversion-based image augmentation approaches are summarized in Fig. \ref{fig:InversionMethods}.

A broad set of methods employ inversion to generate high-quality synthetic data for training other DNNs. For example, the methods of \cite{zhou2023using} and \cite{zhou2023training} generate synthetic images, by inverting real images to the latent space, and then sampling new latent codes around the inverted ones. This facilitates the production of diverse and realistic augmentations of the original dataset, which can significantly improve the performance of downstream classification models.

Another group of methods adopt inversion to enable precise image editing and style transfer. For example, Null-Text Inversion \cite{mokady2023null} consists of two main components, namely pivotal inversion, which estimates an initial diffusion trajectory using DDIM inversion, and `null-text' optimization, which optimizes only the unconditional `null-text' embeddings for each diffusion timestep. Given an input image and its associated text caption, the method inverts the image to obtain a latent code and optimized null-text embeddings. These can then be used to edit the image intuitively using approaches such as P2P \cite{hertz2022prompt}, by modifying the text prompt, while preserving the model weights and conditional embeddings.

Following a similar conceptualization, the method of \cite{zhang2023inversion} enables artistic style transfer from a single reference painting to an input image, using a pretrained SD. In particular, it introduces an attention-based textual inversion module that learns a style representation ("[C]") from the CLIP image embedding of the reference painting, through multi-layer cross-attention. This learnt textual embedding is then encoded into a DM caption conditioning format to guide the generation of new images that combine the content of the input image with the artistic style of the reference painting. Specifically, the method involves extracting the image embedding of the reference painting, learning the corresponding textual embedding via the inversion module, encoding it into DM conditioning format, applying stochastic inversion to the content image to obtain an initial latent noise map, and generating the output image conditioned on the textual embedding and an inverted noise map (noisy version of the input image).

In a closely related way, the method of \cite{li2023stylediffusion} learns to invert the input image to value embeddings. while preserving the keys and attention maps from the original model. The respective architecture consists of a frozen CLIP image encoder, a learnable mapping network, and the SD model. The keys control the output image structure, while the values determine the object style. The editing process involves DDIM inversion to generate latent codes and attention maps, training the mapping network to reconstruct these while preserving object-like attention, and using the trained network with target prompt embeddings for editing.

Following a different line of research, a particular group of methods focus on improving the inversion process itself. For example, EDICT \cite{wallace2023edict} utilizes a coupled DM to perform inversion, where one model learns to map the image to the latent space and the other learns to map the latent code back to the original image space. This allows for more accurate and efficient inversion, compared to optimizing the latent code directly. LocInv \cite{tang2024locinv} utilizes a localization-aware inversion process that optimizes the latent code to match the spatial attention maps of the input image, which facilitates in preserving the local structure and the details of the image during editing. Moreover, the method of \cite{kwon2022diffusion} introduces an asymmetric reverse process (Asyrp) that discovers a semantic latent space, called `h-space', in frozen pretrained DMs. It modifies the RD process by shifting the predicted noise in the bottleneck feature maps of the U-Net architecture, while preserving the direction pointing to the current timestep. This breaks destructive interference and enables semantic manipulation of the generated images. The h-space exhibits desirable properties for editing, such as homogeneity, linearity, composability, robustness, and consistency across timesteps.

\subsubsection{Dataset Expansion}
\begin{figure*}
    \centering
    \includegraphics[width=1.0\linewidth]{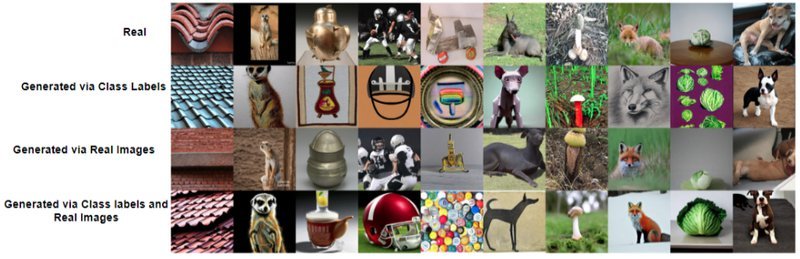}
    \caption{Examples of DM-based dataset expansion image augmentation. Image from \cite{bansal2023leaving}.}
    \label{fig:DatasetExpansion}
\end{figure*}

Conditional DMs are often used for generating synthetic images sampled from the underlying distribution of a given dataset. For example, the method of \cite{zhang2211expanding} trains a class-conditional DM on a small dataset and then uses it to generate additional images per class. The generated examples are filtered using a classifier to ensure that they are realistic and match the target class distribution. Similarly, the approach of \cite{Li2023SemanticGuidedGI} utilizes a semantics-guided DM to generate synthetic images that match the semantic layout and object categories of the original dataset. The method of \cite{wang2022sindiffusion} comprises a single-scale pixel-level DDPM that learns the internal patch distribution of a single natural image. It employs a U-Net denoising network with a restricted patch-level receptive field, enabling it to capture the image patch statistics without memorizing the entire image. The model is trained at a single scale, avoiding error accumulation as observed in progressive growing approaches. The method of \cite{ye2023synthetic} leverages a time-conditioned, U-Net-powered LDM and a two-stage training process to generate realistic synthetic images and to improve classification performance. The first stage involves large-scale pretraining on unlabeled data to learn common features for unconditional image synthesis. The second stage finetunes the model on a small labeled dataset, enabling conditional synthesis guided by a latent classifier. Similarly, the method of \cite{bansal2023leaving} generates synthetic examples that are close to the training data in feature space but far away in the image space, which facilitates towards improving the model's ability to generalize to novel visual concepts (Fig. \ref{fig:DatasetExpansion}).

A different group of methods leverage the ability of DMs to generate high-quality images from text descriptions. For example, TTIDA \cite{yin2023ttida} finetunes a pretrained text-to-text (T2T) model (GLIDE \cite{nichol2021glide}) to generate diverse text descriptions of objects and scenes, and then uses a T2I DM to generate corresponding images. This allows the generation of large-scale synthetic datasets with rich annotations, which can be used to train more accurate and robust vision models.

KNN-Diffusion \cite{sheynin2022knndiffusion} consists of a multi-modal CLIP encoder, a non-trainable retrieval index, and a trainable DM conditioned on the retrieved embeddings. During training, the model receives an image as input, while the CLIP image embedding and its $k$ nearest neighbors are used to condition the generation. During inference, the model receives a text prompt as input, while the CLIP text embedding and its $k$ nearest neighbors are used for conditioning. The DM iteratively denoises a noise vector, guided by the input embedding and the retrieved neighbors, in order to generate the output image. The retrieved neighbors bridge the distribution gap between the image and the text embeddings.

The Retrieval-augmented Diffusion Model (RDM) \cite{blattmann2022semiparametric} combines a trainable conditional DM with a fixed external database of diverse visual examples and a non-trainable retrieval function. During training, the method retrieves from the database the $k$ nearest neighbors for each image, using CLIP embeddings, and encodes them with a pretrained encoder. Subsequently, it conditions the generative decoding head on these encoded representations to generate the target image. During inference, the database and the retrieval function can be flexibly swapped to enable unconditional, class-conditional, text-conditional sampling, or style transfer, by retrieving neighbors based on different criteria or using a database with a different visual style. This approach augments a comparatively small generative model with a large external memory, allowing it to compose novel images based on relevant retrieved information rather than memorizing the full training data; thereby, reducing model size, while enhancing performance and flexibility.

\subsection{Application-Specific Augmentation}
This subsection discusses application-specific DM-powered image augmentation approaches, i.e., methods that take into account particular facts and characteristics that are only present in the examined application domain. Common application fields that have been studied so far are medical imaging, facial recognition, object detection and agriculture, to name a few.

\subsubsection{Medical Imaging}
In the field of medical imaging, image augmentation methods typically employ conditional DMs to generate synthetic medical samples. For example, the methods of \cite{akrout2023diffusion} and \cite{sagers2022improving} train class-conditional DMs on datasets of skin lesion and use them to generate additional examples of each lesion type. The generated images are used to augment the training data and to improve the performance of skin disease classification models. Similarly, the methods of \cite{ali2022spot} and \cite{packhauser2023generation} make use of DMs to generate synthetic chest X-ray images, which are exploited to train more robust models for detecting thoracic abnormalities.

A particular group of methods focus on generating specific types of medical images, such as brain Magnetic Resonance Imaging (MRI) or retinal Optical Coherence Tomography (OCT). For example, the approach of \cite{pinaya2022brain} trains a DM on a dataset of brain MRI images and uses it to generate synthetic images with different neurological conditions. The method of \cite{hu2022unsupervised} trains a DM on retinal OCT images and uses it to generate denoised and super-resolved versions of the images.

A different set of methods employs DMs for image inpainting and anomaly detection. For example, the method of \cite{rouzrokh2022multitask} trains a DM to inpaint brain MRI images with tumors, by conditioning the model on the tumor mask. This allows for the generation of complete and realistic brain images from partial or corrupted scans. In contrast, the approach of \cite{wolleb2022diffusion} employs DDIMs to generate synthetic healthy images from diseased individuals. In particular, a U-Net architecture is trained to iteratively denoise noisy representations, while a binary classifier guides the denoising process towards the healthy class. The method assumes unpaired training data of healthy and diseased images with only image-level labels. For an unseen test image, the DDIM noising process encodes its anatomical information into a noisy representation, which is then iteratively denoised using classifier guidance to generate a corresponding healthy synthetic image. The anomaly map is computed as the pixel-wise difference between the input and the synthetic images, highlighting the diseased regions. The noise level and classifier gradient scale control the trade-off between preserving input details and translating to the healthy class.

\begin{figure*}
    \centering
    \includegraphics[width=1.0\linewidth]{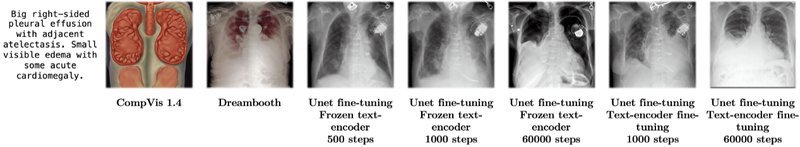}
    \caption{Examples of application-specific DM-based image augmentation (medical imaging), given a specific textual prompt. Image from \cite{chambon2022roentgen}.}
    \label{fig:MedicalImagesExample}
\end{figure*}

Focusing on a different aspect, another group of methods focus on adapting pretrained DMs to the medical domain. For example, the method of \cite{chambon2022adapting} finetunes a pretrained T2I DM on a dataset of chest X-ray images and radiology reports, using textual inversion and cross-attention control. This allows for the generation of synthetic chest X-ray images with specific abnormalities and attributes, based on natural language descriptions. Similarly, the approach of \cite{chambon2022roentgen} finetunes the pretrained SD on a dataset of chest X-rays and radiology reports to adapt it to generate synthetic medical images from text prompts. The key components are a frozen Variatonal Auto-Encoder (VAE) encoder for compressing inputs, a U-Net denoiser conditioned on text embeddings from a CLIP encoder and a VAE decoder. During inference, random noise is encoded and progressively denoised by the U-Net guided by the text prompt, ultimately yielding a synthetic chest X-ray reflecting the described imaging findings (Fig. \ref{fig:MedicalImagesExample}).

More recently, several methods have aimed at further advancing the augmentation performance and quality of the generated images, making use of more recent and sophisticated DM architectures. The method of \cite{guo2023accelerating} introduces PD-DDPM, an accelerated DM for medical image segmentation that uses a pre-segmentation network to generate noisy segmentation predictions, which are then denoised in fewer steps compared to a vanilla DM. The method of \cite{xia2022low} employs a DDPM that is trained on paired low-dose Computed Tomography (CT) and normal-dose CT images. During the FD process, Gaussian noise is gradually added to the normal-dose image based on a time varying schedule, resulting in a noisy image following a standard normal distribution. During the RD process, the U-Net learns to recover the clean normal-dose image from the noisy one, conditioned on the corresponding low-dose image, by predicting and removing the noise at each timestep. To improve sampling efficiency, a fast ordinary differential equation (ODE) solver, termed DPM-Solver \cite{lu2022dpmsolver}, is integrated into the RD process, allowing for faster sampling with fewer steps, while maintaining or improving denoising performance compared to the original DDPM.

\subsubsection{Other Domain-Specific Applications}
Apart from medical imaging, DMs have also been successfully applied to a wide range of domain-specific applications, as detailed in the followings.

\begin{figure*}
    \centering
    \includegraphics[width=0.9\linewidth]{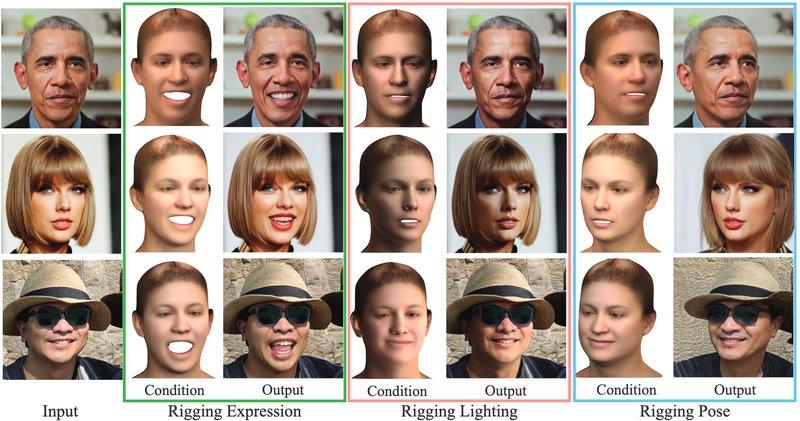}
    \caption{Examples of application-specific DM-based image augmentation (facial recognition and editing), involving different expressions, lighting and poses. Image from \cite{ding2023diffusionrig}.}
    \label{fig:DiffusionRig}
\end{figure*}

\noindent \textbf{Facial Recognition and Editing}: The methods of \cite{boutros2023idiffface} and \cite{huang2024data} focus on generating high-fidelity synthetic faces to improve recognition accuracy. These approaches typically use conditional DMs trained on large-scale facial datasets and generate new faces, by sampling from the learned distribution. Some methods also utilize approaches like multi-level text-related augmentation \cite{wu2023promptrobust} or disentangled representations \cite{ding2023diffusionrig} to enable more fine-grained control over the generated faces, so as to change specific attributes or expressions (Fig. \ref{fig:DiffusionRig}).

\begin{figure*}
    \centering
    \includegraphics[width=1.0\linewidth]{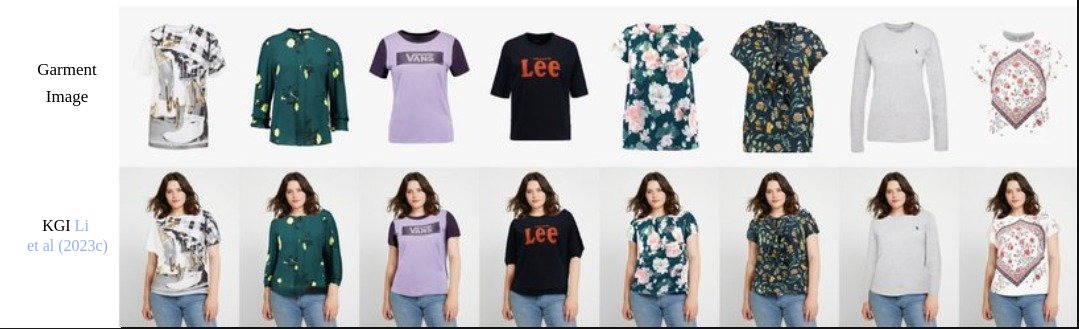}
    \caption{Examples of application-specific DM-based image augmentation (fashion industry), using different input images. Image from \cite{Li_2023_ICCV}}.
    \label{fig:VirtualTryOn}
\end{figure*}

\noindent \textbf{Fashion Industry}: The methods of \cite{Li_2023_ICCV} and \cite{kong2023leveraging} enable realistic virtual try-on experiences, allowing users to visualize clothing items on different body types and poses. These approaches typically use conditional DMs trained on datasets of clothing images and corresponding body poses, and generate new try-on images, by conditioning on the desired clothing item and pose (Fig. \ref{fig:VirtualTryOn}). 

\noindent \textbf{Agriculture} The methods of \cite{deng2023stable}, \cite{harnessing} and \cite{chen2023deep} focus on enhancing datasets for plant disease detection and weed recognition; thereby, improving the accuracy of agricultural models. Such approaches typically employ conditional DMs trained on datasets of plant images with different diseases or weed types, and generate new examples, by sampling from the learned distribution.

\noindent \textbf{Video Editing and Generation}: The methods of \cite{shin2024edit} and \cite{wu2023tuneavideo} enable high-quality video manipulation and generation from text prompts. These approaches typically rely on the use of 3D or video diffusion models that can generate coherent sequences of video frames, based on textual descriptions or input videos.

\noindent \textbf{Cultural Heritage}: The method of \cite{cioni2023diffusion} adopts an LDM to generate diverse, semantically consistent variations of artworks, conditioned on textual descriptions. In particular, the LDM performs a conditional denoising diffusion process in the latent space learned by a convolutional autoencoder, taking the original artwork image and its encoded caption as inputs. By varying the generation seed, the LDM outputs multiple variations of the artwork that preserve the content described in the caption. These synthetic image-caption pairs augment the original art dataset, bridging the domain gap between natural images and artworks, while improving visual grounding of artistic concepts. The augmented dataset, containing both real and synthetic data, facilitates more effective training of downstream vision-language models, like image captioning and cross-modal retrieval ones.

\noindent \textbf{Object Detection}: The approaches of \cite{fang2024data}, \cite{zhang2023diffusionengine}, \cite{li2025simple}, \cite{tang2024aerogen} and \cite{ma2024erase} leverage DMs to generate synthetic data, significantly enhancing model training and performance. These methods make use of conditional DMs trained on object detection datasets and generate new samples, by conditioning on the desired object categories and bounding boxes.

\subsection{Comparison of DM-Based Methods for Image Augmentation}

Having presented the various methodologies for DM-based image augmentation, this section realizes a detailed and in-depth comparison among the most important methods of each category, in order to provide further critical insights. In particular, Table \ref{tab:method_comparison} analyzes several key/important methods, focusing on the following aspects: a) The particular task to be performed, b) If/which FDM is used, c) Input requirements, d) Key exhibited innovation, e) Notable limitations, and f) Fine-tuning requirements. The considered methods, which cover all main categories of the proposed taxonomy (presented in Fig. \ref{fig:data_augmentation} and Table \ref{tab:diffusion_augmentation}) were selected based on the following criteria: a) High citation impact and community adoption, b) Introduction of novel technical approaches or architectures, and c) Demonstration of significant practical capabilities or limitations that influenced subsequent research works.

\begin{sidewaystable*}

\centering
\caption{In-depth comparison of key DM-based augmentation methods.}
\label{tab:method_comparison}
\tiny
\renewcommand{\arraystretch}{0.85}  % Slightly reduce vertical spacing
\setlength{\tabcolsep}{8pt} 

\begin{tabular}{@{}p{1.2cm}p{2cm}p{0.6cm}p{2.5cm}p{3cm}p{3cm}p{2.5cm}@{}}
\midrule[0.03em]
\textbf{Method} & \textbf{Task} & \textbf{FDM} & \textbf{Input Requirements} & \textbf{Key Innovation} & \textbf{Notable Limitations} & \textbf{Fine-tuning Requirements} \\
\midrule \\ 
\multicolumn{7}{c}{\textbf{\footnotesize Semantic Manipulation}} \\ 
\midrule[0.03em]
\cite{yang2023paint} & I2I Translation & SD & Source image + reference image + mask & Content bottleneck prevents trivial copying & May not preserve fine details & Self-supervised training on OpenImages \\
\midrule

\cite{kim2022diffusionclip} & I2I Translation & - & Image + text prompt & Multi-attribute control via noise combination & Limited to pretrained domains & Reference images for domain finetuning \\
\midrule

\cite{lugmayr2022repaint} & Inpainting & - & Image + binary mask & Resampling strategy for mask-agnostic inpainting & Significantly slower than GAN-based methods & - \\
\midrule

\cite{meng2021sdedit} & Editing & - & User strokes or patches & No task-specific training needed & Quality depends on initial noise level & - \\
\midrule

\cite{kawar2023imagic} & Editing & Imagen | SD & Image + text prompt & Linear semantic embedding interpolation & Slow optimization (8min/image) & Target text embedding optimization \\
\midrule

\cite{brooks2023instructpix2pix} & Instruction-based Editing & SD v1.5 & Image + text instruction & Direct instruction followed without optimization & Cannot handle spatial reasoning & Image-instruction-output triplets training \\
\midrule

\cite{Avrahami_2022_CVPR} & Local Editing & - & Image + mask + text prompt & Spatially blended noise guidance for local editing & 30s generation time per image & - \\
\midrule

\cite{chen2024anydoor} & Object Teleportation & SD 2.1 & Target object + scene image + location box & Zero-shot object teleportation with identity preservation & Small detail preservation issues & Light UNet decoder fine-tuning \\
\midrule

\cite{huang2023composer} & Multi-condition Generation & - & Multiple optional conditions (text/sketch/depth/etc.) & Composable conditions with exponential control space & Performance tradeoff in single-condition scenarios & 60M image multi-task training \\
\midrule

\cite{zhang2023adding} & Image Control & SD 1.5/2.1 & Image + text prompt & Zero convolution prevents catastrophic forgetting & Higher GPU memory and training time & Trainable copy with zero convolution \\
\midrule

\multicolumn{7}{c}{\textbf{\footnotesize Personalization and Adaptation}} \\ 
\midrule[0.03em]
\cite{ruiz2023dreambooth} & Personalized Generation & SD & 3-5 concept images + text prompt & Concept-specific generation with few shots & Context-appearance entanglement issues & Class-specific prior preservation loss \\
\midrule

\cite{gal2022image} & Personalized Generation & SD v1.4 & 3-5 concept images & Single word embedding for concept representation & Lengthy optimization (2h per concept) & 5000 optimization steps with reconstruction loss \\
\midrule

\cite{qiu2023controlling} & Personalized Generation & SD 1.5 & 5-50 concept images + text prompt & Orthogonal transformation preserving hyperspherical energy & Color and detail distortions & Orthogonal weight transformation with energy preservation \\
\midrule

\cite{mokady2023null} & Editing & SD v1 & Image + source caption & Null-text optimization without model tuning & Requires editable parts in source caption & - \\
\midrule

\cite{zhang2023inversion} & Style Transfer & SD & Content image + style image & Single-image style learning via attention & Long training time (20min/image) & - \\
\midrule

\cite{wu2023uncovering} & Style Transfer & SD v1.4 & Image + text prompt & Text embedding optimization for attribute editing & Struggles with fine-grained edits & No fine-tuning, only embedding optimization \\
\midrule

\cite{bansal2023leaving} & Classification & SD 1.5 & Class labels or images & Zero-shot data generation for robustness & Limited to ImageNet-like datasets & - \\
\midrule

\cite{zhang2211expanding} & Classification & SD 1.4, DALL-E2 & Small-scale seed dataset & Guided imagination for class-consistent generation & Generated samples less informative than real ones & Domain adaptation fine-tuning \\
\midrule

\multicolumn{7}{c}{\textbf{\footnotesize Application Specific Augmentation}} \\ 
\midrule[0.03em]
\cite{sanchez2022healthy} & Brain Lesion Detection & - & Multi-channel MRI scans & Counterfactual healthy image generation & Limited to brain tumor scenarios & Healthy/unhealthy classification training \\
\midrule

\cite{pinaya2022brain} & Brain MRI Generation & - & Age, sex, brain metrics & Anatomically-controlled brain MRI synthesis & Limited to training distribution range & Conditional training on demographic variables \\
\midrule

\cite{fang2024data} & Object Detection & SD & Visual prior + text prompt & Category-calibrated CLIP filtering & Performance drops with excessive synthetic data & - \\

\bottomrule
\end{tabular}
\end{sidewaystable*}

From the observation of Table \ref{tab:method_comparison}, several key insights can be derived. In particular, SD is shown to be the most widely used FDM across multiple tasks, mainly due to its open-source characteristic. Additionally, an important trade-off between processing speed and output quality is observed, where methods requiring optimization or fine-tuning typically produce better results, but at the cost of longer processing times (often ranging from minutes to hours per image). Another notable trend is the evolution from the usage of single- to multi-condition mechanisms, where methods increasingly support various input modalities (i.e., text, images, masks, sketches) for more precise control; however, this increased flexibility is often accompanied with higher computational requirements and more complex training procedures. Investigating category-specific patterns, semantic manipulation methods typically focus on architectural innovations and generally require less fine-tuning, often implementing zero- or few-shot techniques; these methods also emphasize user control through various input modalities. On the other hand, personalization and adaptation methods demonstrate a clear trend towards optimization-heavy approaches, exhibiting notably longer processing times (often at the scale of hours) and more complex fine-tuning requirements, in order to preserve identity and style characteristics. Moreover, application-specific methods are characterized by their specialized nature, requiring domain-specific training data and often implementing custom architectures or loss functions tailored to their selected application fields; these methods typically prioritize task-specific performance metrics over general-purpose applicability.

\section{Evaluation Metrics}
\label{sec:evaluation}

Evaluating the performance of DM-powered image augmentation methods is crucial to understanding their impact on visual image analysis tasks. This section outlines the methodologies used to assess both the efficacy and efficiency of such frameworks, while a detailed quantitative comparison of DM-based and traditional image augmentation methods is also provided.

\subsection{Quantitative Evaluation}
Quantitative evaluation involves the measurement of the numerical improvements in model performance, as well as the perceptual quality and diversity of the augmented images. This type of evaluation provides objective metrics that can be directly compared and analyzed.
 
\subsubsection{Improvement in Model Performance}
The primary quantitative evaluation involves the measurement of the (potential) improvement in downstream task performance metrics of learned models (e.g., classifiers), trained while incorporating augmented images that have been generated by the DM. The more the downstream model's performance increase, compared to the baseline of realizing training using only the original non-augmented dataset, the better the DM that conducted the augmentation is considered to be. The following cases are most commonly met:
\begin{itemize}
    \item \textbf{Classification Tasks:} Metrics such as accuracy, precision, recall, and F1-score are used to evaluate the performance of classification models. These metrics provide a comprehensive overview regarding how well the model can distinguish between different classes \cite{trabucco2023effective, chowdary2023enhancing, packhauser2023generation, sagers2022improving, bansal2023leaving}.
    \item \textbf{Segmentation Tasks:} Metrics such as Intersection over Union (IoU) and Dice coefficient are used for evaluating segmentation models. IoU measures the overlap between the predicted segmentation and the ground truth one, while the Dice coefficient assesses the similarity between these two \cite{xie2023mosaicfusion, schnell2024scribblegen, sanchez2022healthy, valvano2024controllable}.
    \item \textbf{Object Detection Tasks:} Metrics such as mean Average Precision (mAP), precision and recall are crucial for evaluating object detection models. mAP calculates the average precision across different recall levels and is a standard metric for assessing the accuracy of object detectors. On the other hand, precision and recall provide insights into the detector's ability to correctly identify objects and to minimize false positives \cite{fang2024data, zhang2023diffusionengine}.
\end{itemize}
Analyzing these metrics makes it easier to determine the degree of improvement in the downstream DNN's predictive capabilities, which is introduced by training set augmentation.

\subsubsection{Perceptual Quality and Diversity}
To evaluate the quality and diversity of augmented images, the following metrics are most commonly used:
\begin{itemize}
    \item \textbf{Fr{\'e}chet Inception Distance (FID):} FID \cite{heusel2017gans} measures the distance between the distribution of the generated images and the real ones in the feature space of a pretrained Inception network \cite{szegedy2016rethinking}. It captures both the quality and the diversity of the generated images. Lower FID scores indicate higher quality and more diverse generated images \cite{esser2024scaling, ho2020denoising, rombach2022highresolution, xie2023smartbrush, avrahami2023spatext, gandikota2024unified, zhang2023adding, pinaya2022brain, couairon2022diffedit, fu2024dreamda}.
    \item \textbf{Inception Score (IS):} IS \cite{salimans2016improved} evaluates the quality of the generated images based on the confidence of the class predictions of a pretrained Inception model. High IS values imply that the generated images are diverse and that each image is recognized with high confidence as belonging to a specific class \cite{blattmann2022semiparametric, gafni2020wish, luo2023camdiff, chen2023deep}.
    \item \textbf{Kernel Inception Distance (KID)}: KID \cite{binkowski2018demystifying} comprises an alternative to the case of FID, while it estimates the maximum mean discrepancy between the feature representations of the generated and the real image samples. Lower KID scores indicate better quality and similarity to real images \cite{harnessing, Li_2023_ICCV, kumari2023multi, meng2021sdedit}.
    \item \textbf{Perceptual Metrics}: Metrics such as Learned Perceptual Image Patch Similarity (LPIPS) \cite{zhang2018unreasonable} and Structural Similarity Index Measure (SSIM) \cite{wang2004image} are used to assess the perceptual similarity between generated and real samples, or between the input and the output of image editing tasks \cite{kulikov2023sinddm, wang2022sindiffusion, li2023stylediffusion, zhang2024text, qiu2023controlling, chambon2022adapting, xu2024cyclenet}.
\end{itemize}
Often estimating and combining all above metrics together provides a comprehensive overview of the augmented images' quality and variability.

\subsection{Qualitative Evaluation}
Qualitative evaluation involves subjective assessments of visual realism and relevance of the augmented images. Such types of evaluations are crucial for ensuring that the synthetic data generated by DMs is not only technically sound, but also perceptually convincing and contextually appropriate.

\subsubsection{Visual Quality of Augmented Images}
Assessing the visual quality of augmented images involves inspection by experts, aiming to determine how closely the synthetic images resemble real ones. Expert evaluators are requested to analyze the following main, among others, aspects factors:
\begin{itemize}
    \item \textbf{Realism:} Evaluation of whether the augmented images are indistinguishable from real ones, focusing on characteristics such as texture, lighting, and color consistency \cite{sanchez2022diffusion, kwon2023diffusionbased, dong2022dreamartist}.
    \item \textbf{Detail Preservation:} It is essential to ensure that the augmented images retain critical details necessary for the task, such as fine-grained textures and structural integrity \cite{ruiz2023dreambooth, kawar2023imagic, dong2022dreamartist}.
    \item \textbf{Editing Consistency:} Experts assess how closely the augmented images maintain semantic coherence with the original input and the editing instructions. This involves the evaluation of whether the edits are applied accurately to the intended regions, while preserving the overall context and structure of the image \cite{wang2023instructedit, kawar2023imagic, tang2024locinv, li2023stylediffusion}.
\end{itemize}
Overall, the above described subjective assessment metrics facilitate towards validating the visual authenticity and usability of the generated images.

\subsubsection{Relevance and Contextual Appropriateness}
Evaluating the relevance and contextual appropriateness of augmented images ensures that they maintain semantic coherence and they are suitable for the specific application at hand. This involves the following main, among others, aspects:
\begin{itemize}
    \item \textbf{Contextual Consistency:} Experts review whether the augmented images fit well within the expected context of the task. For example, in medical imaging, synthetic images should accurately reflect the characteristics of the examined disease or the condition being modeled \cite{gandikota2023erasing, han2023highly, ruiz2024hyperdreambooth, wang2023instructedit, zhang2023inversion, tewel2023key}.
    \item \textbf{Semantic Accuracy:} The augmented images should convey the correct semantic information, avoiding any misleading or nonsensical variations. This is critical for applications where accurate representation of objects and scenes is of paramount importance \cite{huang2023region, Li2023SemanticGuidedGI}.
    \item \textbf{Task-Specific Features:} The relevance of augmented images is also evaluated based on their utility for the specific task at hand. For instance, in object detection, the images should contain objects that are correctly annotated and positioned, in order to enhance efficient model training \cite{chambon2022roentgen, valvano2024controllable}.
\end{itemize}
The above described qualitative assessment ensures that the augmented images not only look realistic, but also serve their intended purpose in a satisfactory way.

\subsection{Quantitative Evaluation of DM-based and Traditional Image Augmentation Methods}

\begin{table*}[htbp!]
\caption{Quantitative comparison of DM-based and traditional image augmentation methods. (`Std aug' refers to standard simple geometric transformations, such as rotation, crop, etc.)}
\label{tab:quantitative_comparison}
\tiny

% \definecolor{darkblue}{RGB}{200,245,255}
% \definecolor{lightblue}{RGB}{235,245,255}
% \definecolor{lightgray}{RGB}{222,222,222}

\begin{adjustbox}{width=1.2\textwidth,center}

\begin{tabular}{@{}lllccc@{}}
% \resizebox{\textwidth}{!}{%
% \begin{tabular}{p{1cm} p{0.8cm} p{0.8cm} p{1cm} p{1cm} p{1cm} }
\toprule[\heavyrulewidth]
% \rowcolor{darkblue}
% \toprule
\textbf{Dataset} & \textbf{BASIC-AUG Method} & \textbf{DM Method} & \textbf{BASIC-AUG (\%)} & \textbf{DM-AUG (\%)}  & \textbf{Gain (\%)} \\
\midrule
\multicolumn{6}{c}{\textbf{Classification - Top-1 Accuracy}} \\
\midrule
\multirow{5}{*}{\textit{Aircraft} \cite{maji2013fine}} 
& Guided-SR \cite{kang2023guidedmixup} & \cite{Chen2024DecoupledDA}  & 77.38 & 84.79 & +9.57  \\
\cmidrule{2-6}
 & Cutmix \cite{yun2019cutmix} & \cite{Wang2024EnhanceIC}  & 89.44 & 90.25 & +0.90  \\
\cmidrule{2-6}
 & CAL-AUG \cite{rao2021counterfactual} & \cite{Michaeli2024AdvancingFC} & 81.90 & 87.40 & +6.71  \\
\cmidrule{2-6}
&  GuidedAP \cite{kang2023guidedmixup} & \cite{Islam2024GenMixED} & 84.32 & 85.84 & +1.80  \\
\cmidrule{2-6}
& GuidedAP \cite{kang2023guidedmixup} & \cite{Islam2024DiffusemixLD}  & 84.32 & 85.76 & +1.70  \\
\midrule
\multirow{3}{*}{\textit{Caltech101} \cite{fei2004learning}}
 & RandAugment \cite{cubuk2020randaugment} & \cite{Li2023SemanticGuidedGI} & 58.55 & 59.17 & +1.05   \\
\cmidrule{2-6}
 &  RandAugment \cite{cubuk2020randaugment} & \cite{fu2024dreamda}  & 60.10 & 85.30 & +41.93 \\
\cmidrule{2-6}
 & RandAugment \cite{cubuk2020randaugment} & \cite{zhang2211expanding} & 57.80 & 65.10 & +12.61  \\
\midrule
\multirow{8}{*}{\textit{Cars} \cite{krause20133d}} 
 & RandAugment \cite{cubuk2020randaugment} & \cite{Li2023SemanticGuidedGI} & 86.55 & 88.53 & +2.28  \\
\cmidrule{2-6}
 & RandAugment \cite{cubuk2020randaugment} & \cite{fu2024dreamda} & 65.80 & 87.10 & +32.37   \\
\cmidrule{2-6}
 & RandAugment \cite{cubuk2020randaugment} & \cite{zhang2211expanding} & 43.20 & 75.70 & +75.23  \\
\cmidrule{2-6}
 & Guided-SR \cite{kang2023guidedmixup} & \cite{Chen2024DecoupledDA} & 91.01 & 93.04 & +2.23  \\
\cmidrule{2-6}
 & Cutmix \cite{yun2019cutmix} & \cite{Wang2024EnhanceIC} & 94.73 & 95.21 & +0.50  \\
\cmidrule{2-6}
 & \makecell{RandAugment \cite{cubuk2020randaugment} + \\ Cutmix \cite{yun2019cutmix}} & \cite{Michaeli2024AdvancingFC} & 92.70 & 93.80 & +1.18   \\
\cmidrule{2-6}
 &  GuidedAP \cite{kang2023guidedmixup} & \cite{Islam2024GenMixED} & 90.27 & 91.30 & +1.14  \\
\cmidrule{2-6}
 & GuidedAP \cite{kang2023guidedmixup} & \cite{Islam2024DiffusemixLD} & 90.27 & 91.26 & +1.09  \\
\midrule
\multirow{3}{*}{\textit{CIFAR-100} \cite{krizhevsky2009learning}} 
 &  GuidedAP \cite{kang2023guidedmixup} & \cite{Islam2024DiffusemixLD} & 81.20 & 82.50 & +1.70  \\
\cmidrule{2-6}
 & Cutmix \cite{yun2019cutmix} & \cite{Li2023SemanticGuidedGI} & 77.56 & 75.75 & -2.37   \\
\cmidrule{2-6}
 & GuidedMixup \cite{kang2023guidedmixup} & \cite{Islam2024GenMixED} & 81.20 & 82.58 & +1.69  \\
\midrule
\multirow{3}{*}{\textit{Flowers} \cite{nilsback2008automated}} 
 & RandAugment \cite{cubuk2020randaugment} & \cite{Li2023SemanticGuidedGI} & 41.97 & 45.61 & +8.67  \\
\cmidrule{2-6}
 &  Cutmix \cite{yun2019cutmix} & \cite{zhang2211expanding} & 83.80 & 88.30 & +5.36  \\
\cmidrule{2-6}
 & GuidedMixup \cite{kang2023guidedmixup} & \cite{Wang2024EnhanceIC} & 99.40 & 99.54 & +0.14  \\
\midrule
\multirow{2}{*}{\textit{iWildCam} \cite{koh2021wilds}} 
 & RandAugment \cite{cubuk2020randaugment} & \cite{rahat2024data} & 76.78 & 85.37 & +11.18  \\
\cmidrule{2-6}
 & Cutmix \cite{yun2019cutmix} & \cite{dunlap2023alia} & 77.56 & 84.87 & +9.42  \\
\midrule
\multirow{3}{*}{\textit{CUB} \cite{wah2011caltech}} 
 &  Cutmix \cite{yun2019cutmix} & \cite{rahat2024data} & 70.48 & 73.16 & +3.80  \\
\cmidrule{2-6}
 & Co-Mixup \cite{kim2021co} & \cite{Chen2024DecoupledDA} & 79.41 & 80.82 & +1.77  \\
\cmidrule{2-6}
 & Cutmix \cite{yun2019cutmix} & \cite{Wang2024EnhanceIC} & 87.23 & 87.56 & +0.37  \\
\midrule
\multirow{3}{*}{\textit{Pets} \cite{parkhi2012cats}} 
 & RandAugment \cite{cubuk2020randaugment} & \cite{Li2023SemanticGuidedGI} & 57.45 & 73.71 & +28.39   \\
\cmidrule{2-6}
 &  RandAugment \cite{cubuk2020randaugment} & \cite{fu2024dreamda} & 61.50 & 86.50 & +40.65  \\
\cmidrule{2-6}
  & RandAugment \cite{cubuk2020randaugment} & \cite{zhang2211expanding} & 48.00 & 73.40 & +52.91 \\
\midrule
\multicolumn{6}{c}{\textbf{Few-shot Classification (8 examples per class) - Top-1 Accuracy}} \\
\midrule
\multirow{2}{*}{\textit{MS COCO} \cite{lin2014microsoft}} 
 & Std aug & \cite{trabucco2023effective} & 42.00 & 47.00 & +11.90 \\
\cmidrule{2-6}
 & Std aug  & \cite{lingenberg2024diagen} & 47.00 & 57.00 & +21.67 \\
\midrule
\multicolumn{6}{c}{\textbf{Object Detection - mAP}} \\
\midrule
\multirow{1}{*}{\textit{PASCAL VOC} \cite{everingham2010pascal}} 
 & RandAugment \cite{cubuk2020randaugment} & \cite{li2025simple} & 78.20 & 79.10 & +1.15 \\
\midrule
\multirow{1}{*}{\textit{DIOR-R} \cite{cheng2022anchor}} 
 & CopyPaste + Flip \cite{dwibedi2017cut} & \cite{tang2024aerogen} & 38.75 & 41.69 & +7.58  \\
\midrule
\multicolumn{6}{c}{\textbf{Out-of-Distribution Classification - OOD Accuracy}} \\
\midrule
\multirow{3}{*}{\textit{Waterbirds} \cite{sagawa2019distributionally}} 
 & RandAugment \cite{cubuk2020randaugment} & \cite{dunlap2023alia} & 30.32 & 46.63 & +53.79  \\
\cmidrule{2-6}
 & Mixup \cite{zhang2017mixup} & \cite{Chen2024DecoupledDA} & 72.52 & 76.16 & +5.03 \\
\cmidrule{2-6}
 & Cutmix \cite{yun2019cutmix} & \cite{Wang2024EnhanceIC} & 71.23 & 72.47 & +1.74 \\
\midrule
\multicolumn{6}{c}{\textbf{Segmentation - mIoU}} \\
\midrule
\multirow{3}{*}{\textit{KITTI} \cite{fritsch2013new}} 
 & Cutout \cite{devries2017improved} & \cite{ma2024erase} & 78.65 & 82.20 & +4.51 \\
\cmidrule{2-6}
 & Cutout \cite{devries2017improved} & \cite{ma2024erase} & 77.65 & 78.34 & +0.88 \\
\cmidrule{2-6}
 & Cutout \cite{devries2017improved} & \cite{ma2024erase} & 92.35 & 92.72 & +0.40 \\
\midrule
\multirow{1}{*}{\textit{PASCAL VOC} \cite{everingham2010pascal}} 
 & Std aug & \cite{schnell2024scribblegen} & 78.10 & 78.90 & +1.02 \\
\midrule
\multicolumn{6}{c}{\textbf{Medical Image Classification - Top-1 Accuracy}} \\
\midrule
\multirow{1}{*}{\textit{OrganMNIST} \cite{yang2021medmnist}} 
 & RandAugment \cite{cubuk2020randaugment} & \cite{zhang2211expanding} & 79.60 & 80.70 & +1.38 \\
\midrule
\multirow{1}{*}{\textit{PathMNIST} \cite{yang2021medmnist}} 
 & RandAugment \cite{cubuk2020randaugment} & \cite{zhang2211expanding} & 79.20 & 86.90 & +9.72 \\
\midrule
\multirow{1}{*}{\textit{BreastMNIST} \cite{yang2021medmnist}} 
 & RandAugment \cite{cubuk2020randaugment} & \cite{zhang2211expanding} & 68.70 & 77.40 & +12.66 \\
\midrule
\multirow{1}{*}{\textit{Shenzhen TB} \cite{jaeger2014two}} 
& RandAugment \cite{cubuk2020randaugment} & \cite{fu2024dreamda}  & 75.50 & 83.50 & +10.59 
\\
\bottomrule
\end{tabular}
\end{adjustbox}
\end{table*}

Having presented the various quantitative and qualitative evaluation metrics, this subsection realizes a comprehensive quantitative comparison among DM-based and traditional image augmentation methods. In particular, Table \ref{tab:quantitative_comparison} reports the performance merits of traditional (namely the approaches of \cite{kang2023guidedmixup, yun2019cutmix, rao2021counterfactual, cubuk2020randaugment, kim2021co, dwibedi2017cut, zhang2017mixup, devries2017improved}) and key DM-based augmentation methods, applied separately to common downstream tasks, namely image classification, object detection and semantic segmentation. The table illustrates the reported improvement that each DM-based augmentation method (denoted DM-AUG) induces, compared to the considered traditional one (denoted BASIC-AUG).

From the observation of Table \ref{tab:quantitative_comparison}, it can be seen that almost all DM-based augmentation methods demonstrate performance improvements across a wide range of tasks and datasets, over conventional augmentation approaches. The improvement level varies across different tasks, exhibiting substantial (or outstanding) gains in some cases and more modest enhancements in others. It needs to be highlighted though that this increase in performance is also associated with correspondingly significant computational overhead for DM-based methods. In particular, DM-powered generation time may vary considerably across methods and hardware setups, ranging between 0.43-6.6 seconds per image using high-performing GPUs (e.g., H800, V100, RTX3090, etc.), while traditional methods (e.g., Cutout, GridMask, etc.) operate in approximately 0.008 seconds per image \cite{zhang2211expanding, fu2024dreamda}. Indicatively, the total generation time of DM-based methods for whole datasets may vary for relatively small datasets from 2 hours (e.g., Watebirds \cite{sagawa2019distributionally}) up to 7 hours (e.g., iWildCam \cite{koh2021wilds}), while certain methods may require up to 10 hours to generate 100K images \cite{dunlap2024diversify, du2024dream}. Apart from computational cost, DM-powered methods also often require additional storage space for maintaining the generated images, unlike traditional augmentation approaches that can be performed on-the-fly \cite{Islam2024GenMixED}. Moreover, prompt quality dependency is usually a common bottleneck that requires proper construction and careful selection; poor prompt selection could lead to unrealistic or unsuitable generated images that may degrade performance \cite{Islam2024DiffusemixLD, Islam2024GenMixED}.

\section{Challenges and Future Research Directions}
\label{sec:limitations}

Image augmentation using DM-powered methodologies constitutes a promising approach for enhancing the diversity and quality of training datasets. However, despite the rapid progress and significant advances that have been observed in the field in the recent period, several open challenges, which at the same time comprise future research direction, remain to be addressed. These can broadly be divided into general ones, which concern the application of DMs in general and regardless of the particular application case, and challenges that are particularly important for image augmentation.

\subsection{Computational Cost and Efficiency}
A significant factor associated with the application of any DM-based architecture concerns its very high computational cost. In particular, DM require substantial computational resources and can be time-consuming for both training and inference, which may delay their development and deployment. On the contrary, a trained Generative Adversarial Network (GAN) Generator synthesizes an output image directly, i.e., with just one forward pass. The iterative denoising process of DMs renders it relatively difficult to scale such methods to real-world applications, large datasets and complex tasks. For example, SD requires more than 200,000 GPU hours \cite{rombach2022highresolution} to be trained on the LAION-5B dataset with one A100 40GB GPU. Approaches such as DDIM and DPM-Solver++ \cite{lu2022dpm} target to speed up the sampling process and to improve efficiency. 
% However, further research is needed to develop more efficient architectures and sampling methods that can generate high-quality images with reduced computational overhead.

Given this context, a significant avenue for future research comprises the development of more efficient DM architectures and optimization methods. Promising directions include knowledge distillation approaches \cite{luo2023comprehensive} to create lightweight models \cite{song2024lightweight}, quantization \cite{shang2023post} or advanced parallelization strategies \cite{shih2024parallel}. Recent work on progressive distillation \cite{salimans2022progressive} and one-step generation \cite{yin2024one} shows potentials for dramatically reducing inference time. Moreover, research into specialized acceleration \cite{ma2024deepcache, wang2024patch} could facilitate towards bridging the efficiency gap between DMs and other generative approaches.

\subsection{Lack of Fine-grained Control and Interpretability}
Interpretability and controllability of DMs is generally problematic, making it challenging to understand how they generate their outputs (e.g., given specific conditions). The users have limited ability to precisely control specific attributes, objects, or regions in the generated images, which can hinder the usefulness of these methods for certain applications. 
% Developing methods for interpreting the output of DM-based models is an important area of research. Approaches such as classifier-free guidance, cross-attention control or spatial conditions (e.g., masks, regions etc.) have been proposed to mitigate these issues and to enable control of the sampling process. However, further research is required in this area.

As a result, it is important for future research to explore advanced conditioning mechanisms and interpretability tools. Research directions may include developing more sophisticated attention-based control mechanisms \cite{hertz2022prompt} and creating interactive interfaces for fine-grained manipulation \cite{lee2024semanticdrawrealtimeinteractivecontent}.

\subsection{Limited Diversity and Realism of Generated Data}
One of the most prevalent challenges is the (relatively) constrained diversity and realism of the synthetic images generated by DMs. While DMs have shown impressive capabilities in generating high-quality images, they often struggle to capture the full diversity and complexity of real-world data distributions. This limitation can lead to a domain gap between the synthetic and real data, which can hinder the effectiveness of using the generated data for training downstream tasks, like image classification and object detection. 

Several methods have been introduced so far that aim to address this issue by incorporating approaches that improve the diversity and realism of the generated samples, such as using language enhancement \cite{dunlap2022using} and post-filtering (e.g., CLIP-based filtering \cite{kim2022diffusionclip}). Further relevant research towards the direction of domain adaptation \cite{chopra2024source} or style transfer \cite{kwon2023diffusionbased} may facilitate towards bridging the gap between real and synthetic image distributions.
% However, more intense and in-depth research is needed to develop more advanced methods that can generate truly diverse and realistic synthetic data across various domains.

\subsection{Model Overfitting and Catastrophic Forgetting}
One of the most common challenges faced by many of the methods covered in this study comprises the issue of model overfitting and catastrophic forgetting \cite{Cywinski2024GUIDEGI}. In particular, overfitting occurs when a model learns to fit the training data too closely, at the expense of generalization to new, unseen data. This can be especially problematic when working with limited training data, as is often the case in domains like medical imaging. On the other hand, catastrophic forgetting refers to the tendency of trainable models to forget previously learned representations, when being finetuned on new data or tasks. This is a major hurdle for approaches that aim to adapt pretrained models to specific domains or styles. Directly finetuning large models on limited datasets can lead to a rapid loss of the originally learned knowledge structures. 

Recent approaches that target to mitigate this \cite{zeng2024infusion,zhong2024diffusion,zajkac2023exploring} leverage various methodologies, including selective parameter updates, dual-stream architectures, and careful management of the denoising process, to maintain model generalization while adapting to new tasks or concepts. Future research along the promising directions of relevant continual learning \cite{smith2023continual} or metalearning \cite{zhang2024metadiff} approaches may bear the potentials for further alleviating any remaining issues.

\subsection{Evaluation Metrics and Benchmarks}
Evaluating the effectiveness of DM-powered image augmentation remains a challenge itself. Traditional metrics like FID and KID may not fully assess/capture the quality and diversity of the generated samples, especially for complex or specialized domains. Additionally, the lack of standardized benchmarks and evaluation protocols makes it difficult to compare different methods and to assess their generalization and augmentation abilities. 

To this end, developing more comprehensive and both general as well as domain-specific evaluation metrics, while also establishing common benchmarks and datasets, comprises an important area for future research for enabling more rigorous and consistent evaluation \cite{betzalel2024evaluation}.

\subsection{Ethical Considerations and Bias}
Large-scale T2I DMs are often trained on Web-scraped datasets, which can contain harmful stereotypes, offensive content, and biases related to gender, race, age, and other sensitive attributes. These biases can get amplified in the generated images, leading to unfair representations and perpetuating societal stereotypes. An additional ethical consideration is the use of copyrighted or private data in the training datasets without proper consent or attribution. This raises significant concerns regarding the ownership and fair use of the generated images, as well as the potential for privacy violations.

Such concerns may partially be mitigated by technical measures, which, however, would require research on detecting and mitigating biases in both training and generated images. Similarly, creating fairness-aware training procedures \cite{friedrich2023fair} and establishing ethical guidelines for data collection and model deployment \cite{zhang2023text} are potentially worthy avenues for investigation, towards more transparent and trustworthy DMs.

\section{Conclusion}
\label{sec:conclusion}

Image augmentation comprises a fundamental task in modern computer vision, since it allows the enhancement of training datasets with realistic synthetic samples, permitting the automatic context- and semantics-aware editing of given reference images, etc. Diffusion Models (DMs) have shown significant promise in generating realistic and diverse images, capturing complex relationships and structures in high-dimensional image data. Moreover, the ability to condition the generation process using class labels, textual descriptions, or visual prompts allows for targeted augmentation, generating images that fulfill specific requirements based on the task at hand.

This study provided a comprehensive overview of the recent advancements in DM-powered image augmentation, a taxonomy of the main categories of existing methods, insights regarding the practical usage of DM-powered techniques for semantic manipulation, personalization and adaptation, and application-specific image augmentation, and a review of the relevant performance evaluation metrics. Promising and significant avenues for future research include the improvement of the efficiency of DMs, i.e., reducing computational costs and improving scalability, enhancing interpretability and control over generated images, as well as boosting diversity and realism of synthetic data. Finally, developing new, robust evaluation metrics and addressing ethical considerations emerges as a critical key to the progress and responsible deployment of DMs in image augmentation.

\bmhead{Acknowledgments}
The research leading to the results of this paper has received funding from the European Union’s Horizon Europe research and development programme under grant agreement No. 101073876 (Ceasefire).

\section*{Declarations}
\textbf{Funding} The research leading to these results received funding from the European Commission under Grant Agreement No. 101073876 (Ceasefire).\\\\
\textbf{Conflict of interest} The authors have no competing interests to declare that are relevant to the content of this article.\\\\
\textbf{Author contribution} Panagiotis Alimisis, Ioannis Mademlis and Georgios Th. Papadopoulos performed the literature review and prepared a draft of the manuscript. Panagiotis Radoglou-Grammatikis and Panagiotis Sarigiannidis performed reviewing and editing of the manuscript. Georgios Th. Papadopoulos was responsible for receiving the funding to implement the study.

\bibliography{references}% common bib file

\end{document}